\documentclass[10pt,twocolumn,letterpaper]{article}

\usepackage[pagenumbers]{cvpr} 

\usepackage{times}
\usepackage{epsfig}
\usepackage{graphicx}
\usepackage{amsmath}
\usepackage{amssymb}
\usepackage{multirow}
\usepackage{graphics}
\usepackage{threeparttable}
\usepackage{color}
\usepackage[normalem]{ulem}
\usepackage{float}
\usepackage{amsfonts}
\usepackage{bm}
\usepackage{bbm}
\usepackage{enumitem}
\usepackage[numbers]{natbib}
\usepackage{subcaption}
\usepackage{algorithm}
\usepackage{array}
\usepackage[table]{xcolor}
\usepackage{colortbl}
\usepackage{pifont}

\usepackage{mathtools}
\usepackage{siunitx}
\usepackage[nopar]{lipsum}
\usepackage{listings}
\usepackage[export]{adjustbox}

\usepackage{mathtools}
\usepackage{siunitx}
\usepackage[nopar]{lipsum}
\usepackage{listings}
\usepackage{etoolbox}
\makeatletter
\AfterEndEnvironment{algorithm}{\let\@algcomment\relax}
\AtEndEnvironment{algorithm}{\kern2pt\hrule\relax\vskip3pt\@algcomment}
\let\@algcomment\relax
\newcommand\algcomment[1]{\def\@algcomment{\footnotesize#1}}
\renewcommand\fs@ruled{\def\@fs@cfont{\bfseries}\let\@fs@capt\floatc@ruled
	\def\@fs@pre{\hrule height.8pt depth0pt \kern2pt}%
	\def\@fs@post{}%
	\def\@fs@mid{\kern2pt\hrule\kern2pt}%
	\let\@fs@iftopcapt\iftrue}
\makeatother

\newcolumntype{I}{!{\vrule width 1pt}}

\definecolor{myy}{RGB}{126,95,0}
\definecolor{mygray}{gray}{.9}
\definecolor{bblue}{RGB}{30,80,120}
\definecolor{mygray1}{gray}{.7}
\definecolor{ggray}{RGB}{127,127,127}

\newcommand{\cmark}{\ding{51}}%
\newcommand{\pub}[1]{{\color{gray}{\tiny{[{#1}]}}}}

\def\1{\mathbbm{1}}
\makeatletter\renewcommand\paragraph{\@startsection{paragraph}{4}{\z@}
	{.5em \@plus1ex \@minus.2ex}{-.5em}{\normalfont\normalsize\bfseries}}\makeatother

\definecolor{mygreen}{HTML}{39b54a}  

\makeatletter
\newcommand{\thickhline}{%
	\noalign {\ifnum 0=`}\fi \hrule height 1pt
	\futurelet \reserved@a \@xhline
}
\makeatother

\usepackage[pagebackref=true,breaklinks=true,letterpaper=true,colorlinks,bookmarks=false]{hyperref}

\usepackage{tabulary}
\newcolumntype{I}{!{\vrule width 1pt}}
\newcommand{\tabincell}[2]{\begin{tabular}{@{}#1@{}}#2\end{tabular}}
\newcolumntype{x}[1]{>{\centering\arraybackslash}p{#1pt}}
\newcolumntype{y}[1]{>{\raggedright\arraybackslash}p{#1pt}}
\newcolumntype{z}[1]{>{\raggedleft\arraybackslash}p{#1pt}}
\newlength\savewidth
\newcommand{\tablestyle}[2]{\setlength{\tabcolsep}{#1}\renewcommand{\arraystretch}{#2}\centering\footnotesize}
\newcommand{\myhyperlink}[3][black]{\hyperlink{#2}{\color{#1}{#3}}}
\usepackage[capitalize]{cleveref}
\crefname{section}{§}{§§}
\Crefname{section}{§}{§§}

\def\cvprPaperID{3376} 
\def\confName{CVPR}
\def\confYear{2022}

\begin{document}

\title{Rethinking Semantic Segmentation: A Prototype View}

\author{Tianfei Zhou$^1$,~~Wenguan Wang$^{2,1}$\thanks{Corresponding author: \textit{Wenguan Wang}.}~,~~Ender Konukoglu$^1$,~~Luc Van Gool$^1$   \\
  \small{$^1$ Computer Vision Lab, ETH Zurich} \hspace{0pt} \small{$^2$ ReLER, AAII, University of Technology Sydney} \\
  \small\url{https://github.com/tfzhou/ProtoSeg}
}
\maketitle

\begin{abstract}
Prevalent$_{\!}$ semantic$_{\!}$ segmentation$_{\!}$ solutions,$_{\!}$ despite$_{\!}$ their different network designs (FCN based or attention based) and$_{\!}$ mask$_{\!}$ decoding$_{\!}$ strategies$_{\!}$ (parametric$_{\!}$ softmax$_{\!}$ based$_{\!}$ or pixel-query based), can be placed in one category, by  con- sidering the softmax weights or query vectors as learnable\\
\noindent class$_{\!}$ prototypes.$_{\!}$ In$_{\!}$ light$_{\!}$ of$_{\!}$ this$_{\!}$ prototype$_{\!}$ view,$_{\!}$ this$_{\!}$~study$_{\!}$~un- covers several limitations of such parametric~segmentation\\
\noindent regime, and proposes a nonparametric alternative based on\\
\noindent non-learnable$_{\!}$ prototypes.$_{\!}$ Instead$_{\!}$ of$_{\!}$  prior$_{\!}$~methods$_{\!}$ learning a single weight/query vector  for each~class in a~fully parametric manner, our model represents each class as a set of\\
\noindent  non-learnable prototypes, relying solely~on the mean fea- tures of several training pixels within that class. The dense\\
\noindent prediction is thus achieved by nonparametric nearest prototype retrieving. This  allows our model to directly shape the  pixel embedding space,  by optimizing the arrangement be- tween$_{\!}$ embedded$_{\!}$ pixels$_{\!}$ and$_{\!}$ anchored$_{\!}$ prototypes.$_{\!}$ It$_{\!}$ is$_{\!}$ able$_{\!}$~to
handle
arbitrary number of classes with a constant amount of$_{\!}$ learnable$_{\!}$ parameters.$_{\!\!}$
We$_{\!}$ empirically$_{\!}$ show$_{\!}$ that,$_{\!}$ with$_{\!}$~FCN$_{\!}$ based and attention based segmentation models (\ie, HR- Net,${\!}$ Swin,${\!}$ SegFormer)${\!}$ and${\!}$ backbones${\!}$ (\ie,${\!}$ ResNet,${\!}$ HRNet, Swin, MiT), our nonparametric framework yields compel- ling results over several datasets (\ie, ADE20K,~Cityscapes,

\noindent COCO-Stuff), and performs well in the large-vocabulary situation. We expect this work will provoke a rethink of the current de facto semantic segmentation model design.
\end{abstract}
\vspace{-10pt}
\section{Introduction}\label{sec:intro}
\vspace{-1pt}



%

\begin{figure}[t]
	\vspace{-10pt}
	\begin{center}
		\includegraphics[width=\linewidth]{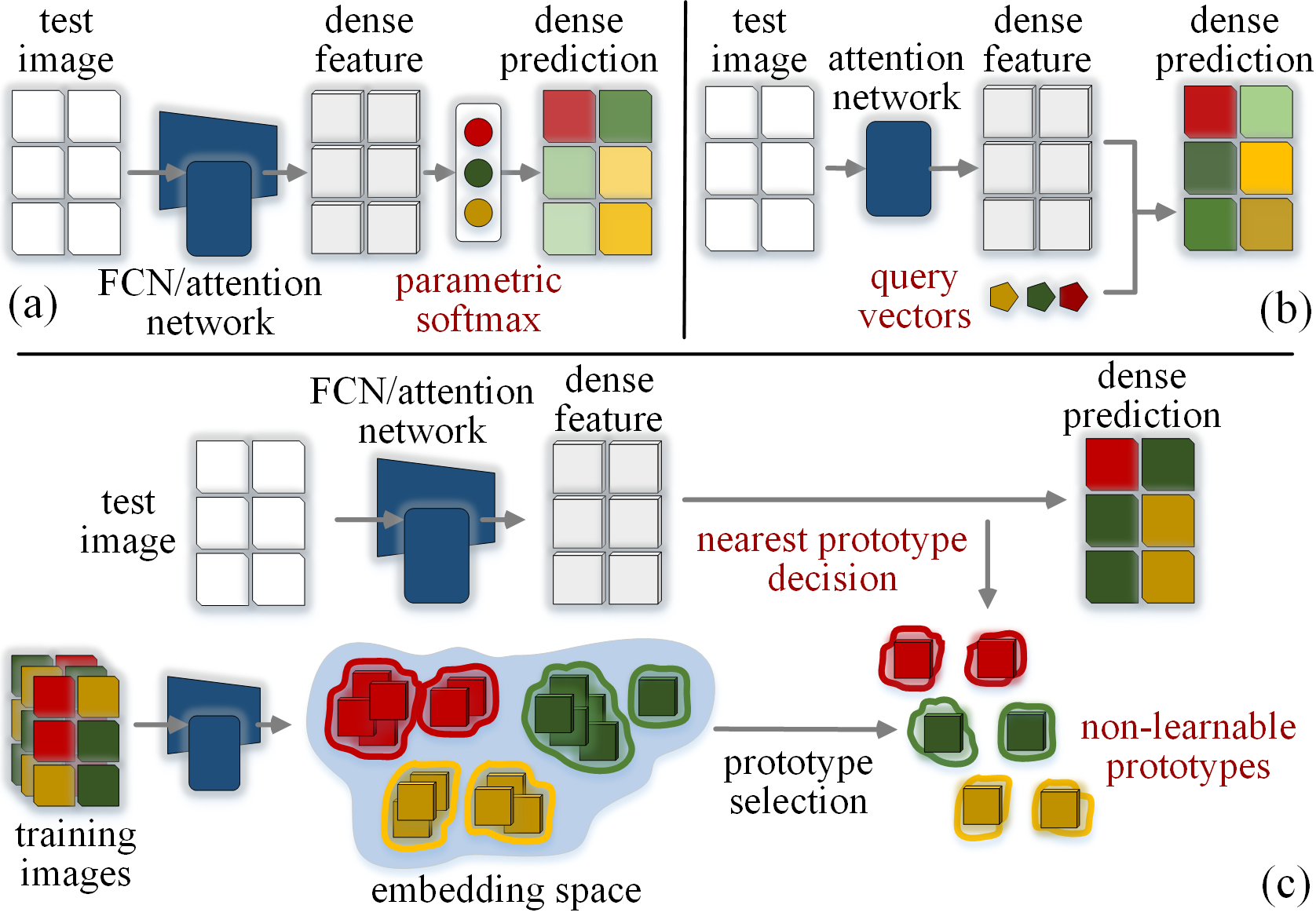}
	\end{center}
	\vspace{-18pt}
	\captionsetup{font=small}
	\caption{\small\!{Different sematic segmentation paradigms}: (a-b) \textbf{parametric}$_{\!}$ \textit{vs}$_{\!}$ (c)$_{\!}$ \textbf{nonparametric}.$_{\!}$ Modern$_{\!}$ segmentation$_{\!}$ solutions,$_{\!}$ no matter using (a) parametric softmax or (b) query vectors for mask decoding, can be viewed as learnable prototype based methods that learn class-wise prototypes in a fully parametric manner.  We~in- stead propose a nonparametric scheme (c) that directly selects subcluster centers of embedded pixels as prototypes, and achieves per-pixel prediction via nonparametric nearest prototype retrieving. }
	\vspace{-15pt}
	\label{fig:motivation}
\end{figure}

With the renaissance of connectionism, rapid progress has been made in semantic segmentation. Till now, most~of state-of-the-art segmentation models~\cite{chen2017deeplab,zhao2017pyramid,hu2018squeeze,fu2019dual} were built upon {Fully Convolutional Networks} (FCNs)$_{\!}$~\cite{long2015fully}. Despite their diversified model designs and impressive results, existing \textit{FCN based} methods commonly apply \textit{\textbf{parametric softmax}} (\includegraphics[scale=0.6,valign=c]{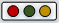}) over pixel-wise features for dense prediction (Fig.$_{\!}$~\ref{fig:motivation}(a)).$_{\!}$ Very$_{\!}$ recently,$_{\!}$ the$_{\!}$ vast success of Transformer \cite{vaswani2017attention} stimulates the emergence of \textit{attention based} segmentation solutions. Many$_{\!}$ of$_{\!}$ these$_{\!}$ `non-FCN'$_{\!}$ models,$_{\!}$ like \cite{zheng2021rethinking,xie2021segformer}, directly follow the standard mask decoding
regime,$_{\!}$ \ie,$_{\!}$ estimate$_{\!}$ softmax$_{\!}$ distributions$_{\!}$ over$_{\!}$ dense$_{\!}$ visual$_{\!}$
embeddings (extracted from patch token sequences). Interestingly, the others \cite{strudel2021segmenter,cheng2021maskformer} follow the good practice of Transformer in other fields \cite{carion2020end,wang2021end,meinhardt2021trackformer} and  adopt a \textit{\textbf{pixel-query}} strategy (Fig.$_{\!}$~\ref{fig:motivation}(b)): utilize a set of learnable vectors  (\includegraphics[scale=0.7,valign=c]{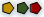}) to query the dense embeddings for mask prediction. They speculate the learned query vectors can capture class-wise properties, however, lacking in-depth analysis.

Noticing there exist two different mask decoding strategies, the following questions naturally arise: \hypertarget{Q1}{\ding{182}}
\textit{What are the relation and difference  between them?} and \hypertarget{Q2}{\ding{183}}  \textit{If the learnable query vectors indeed implicitly capture some intrinsic properties of data, is there any better way to achieve this?}

Tackling these two issues can provide insights~into modern segmentation model design, and motivate us to rethink the task from \textit{a} \textit{prototype} \textit{view}. The idea of prototype based classification$_{\!}$~\cite{duda1973pattern} is classical and intuitive (which can date back to the$_{\!}$ nearest$_{\!}$ neighbors$_{\!}$ algorithm$_{\!}$~\cite{cover1967nearest}$_{\!}$ and$_{\!}$ find$_{\!}$~evi-dence in cognitive science$_{\!}$~\cite{rosch1973natural,knowlton1993learning}):$_{\!}$ data samples are classi- fied based on their proximity to \textit{representative} prototypes~of classes. With this perspective, in \S\ref{sec:rev}, we first answer question \myhyperlink{Q1}{\ding{182}} by pointing out most modern segmentation methods, from softmax based to pixel-query based, from FCN based to attention based, fall into one grand category: \textit{\textbf{parametric}} models$_{\!}$ based$_{\!}$ on \textit{\textbf{learnable$_{\!}$ prototypes}}.$_{\!}$ Consider$_{\!}$ a$_{\!}$ segmenta-$_{\!}$ tion$_{\!}$ task$_{\!}$ with$_{\!}$
$C$$_{\!}$ semantic$_{\!}$ classes.$_{\!}$ Most$_{\!}$ existing$_{\!}$~efforts
seek to directly \textit{learn} $C$ class-wise prototypes -- softmax~weights or query vectors -- for parametric, pixel-wise classification.  Hence  question  \myhyperlink{Q2}{\ding{183}} becomes more fundamental:$_{\!}$ \hypertarget{Q3}{\ding{184}}$_{\!}$  \textit{What are the limitations of this learnable  prototype  based parametric paradigm?}  and \hypertarget{Q4}{\ding{185}}$_{\!}$  \textit{How to  address these  limitations?}


$_{\!}$Driven$_{\!}$ by$_{\!}$ question$_{\!}$ \myhyperlink{Q3}{\ding{184}},$_{\!}$ we$_{\!}$ find$_{\!}$ there$_{\!}$ are$_{\!}$ three$_{\!}$ critical$_{\!}$ limi-\\
\noindent  tations:$_{\!}$ \textit{\textbf{First}},$_{\!}$ usually$_{\!}$ only$_{\!}$ one$_{\!}$ single$_{\!}$ prototype$_{\!}$ is$_{\!}$ learned$_{\!}$~per class, insufficient to describe rich intra-class variance. The prototypes are simply learned in a fully parametric manner, without considering their representative ability. \textit{\textbf{Second}}, to map a $H\!\times\!W_{\!}\times_{\!}D_{\!}$ image
feature tensor into a $H\!\times\!W_{\!}\times_{\!}C_{\!}$~semantic mask, at$_{\!}$ least $D\!\times\!C$
parameters$_{\!}$ are needed for prototype\\
\noindent  learning.$_{\!}$ This$_{\!}$ hurts$_{\!}$  generalizability$_{\!}$~\cite{wu2018improving},$_{\!}$ especially$_{\!}$ in$_{\!}$~the large-vocabulary case; for instance, if there are $800$ classes  and$_{\!}$  $D_{\!}\!=_{\!}\!512$,$_{\!}$ we$_{\!}$
need 0.4M$_{\!}$ learnable$_{\!}$ prototype$_{\!}$ parameters
\textit{alone}.$_{\!}$ \textit{\textbf{Third}},$_{\!}$ with$_{\!}$ the$_{\!}$ cross-entropy$_{\!}$ loss,$_{\!}$ only the relative relations between$_{\!}$ intra-class$_{\!}$ and$_{\!}$ inter-class$_{\!}$ distances$_{\!}$ are$_{\!}$ optimized \cite{zhang2020rbf,pang2020rethinking,wang2021exploring}; the actual distances between pixels and prototypes, \ie, intra-class compactness, are ignored$_{\!}$. 


As$_{\!}$ a$_{\!}$ response$_{\!}$ to$_{\!}$ question$_{\!}$~\myhyperlink{Q4}{\ding{185}},$_{\!}$ in$_{\!}$~\S\ref{sec:method},$_{\!}$ we$_{\!}$ develop$_{\!}$ a$_{\!}$ \textit{\textbf{nonpa- rametric}} segmentation framework, based on \textit{\textbf{non-learnable prototypes}}.$_{\!\!}$ Specifically,$_{\!}$ building$_{\!}$ upon$_{\!}$ the$_{\!}$ ideas$_{\!}$ of$_{\!}$ prototype\\
\noindent learning$_{\!}$~\cite{zhang2009prototype,wu2018unsupervised}$_{\!}$ and$_{\!}$ metric$_{\!}$ learning$_{\!}$~\cite{hadsell2006dimensionality,li2020prototypical},$_{\!}$ it$_{\!}$ is$_{\!}$~fully aware~of the limitations of its parametric counterpart. Inde- pendent$_{\!}$ of$_{\!}$ specific$_{\!}$ backbone$_{\!}$ architectures$_{\!}$ (FCN based or attention based), our method is general and brings insights into segmentation model design and training. For model design, our method explicitly sets \textit{sub-class centers}, in the pixel embedding space, as the prototypes. Each pixel data is predicted$_{\!}$ to$_{\!}$ be$_{\!}$ in$_{\!}$ the$_{\!}$ same$_{\!}$ class$_{\!}$ as$_{\!}$ the$_{\!}$ nearest$_{\!}$ prototype, {without relying on extra learnable parameters}. For training, as the prototypes are representative of the dataset, we can directly pose known inductive biases (\eg, intra-class compactness, inter-class separation) as extra optimization criteria and efficiently shape the whole embedding space, instead$_{\!}$ of$_{\!}$ optimizing$_{\!}$ the$_{\!}$ prediction$_{\!}$ accuracy$_{\!}$ only.$_{\!}$ Our
model has three$_{\!}$ appealing$_{\!}$ advantages:$_{\!}$ \textit{\textbf{First}},$_{\!}$ each$_{\!}$ class$_{\!}$ is$_{\!}$ abstracted$_{\!}$ by$_{\!}$
a set of prototypes, well capturing class-wise characteristics and intra-class variance. With the clear meaning of the prototypes, the interpretability is also enhanced -- the prediction of each pixel can be intuitively understood as the reference of its closest class center in the embedding space~\cite{biehl2009metric,backhaus2014classification}. \textit{\textbf{Second}}, due to the nonparametric nature, the generalizability is improved. Large-vocabulary semantic segmentation can also be handled efficiently, as the amount of learnable prototype parameters is no longer constrained to the number of classes (\ie, 0 \textit{vs} $D\!\times\!C$). \textit{\textbf{Third}},
via prototype-anchored metric$_{\!}$ learning,$_{\!}$ the$_{\!}$ pixel$_{\!}$ embedding$_{\!}$ space$_{\!}$ is$_{\!}$ shaped$_{\!}$ as$_{\!}$ well-structured, benefiting segmentation prediction eventually.

$_{\!}$By$_{\!}$ answering$_{\!}$ questions$_{\!}$ \myhyperlink{Q1}{\ding{182}}-\myhyperlink{Q4}{\ding{185}},$_{\!}$ we$_{\!}$ formalize$_{\!}$ prior$_{\!}$ methods\\
\noindent  within a learnable prototype based, parametric framework, and link this field to prototype learning and metric learning. We$_{\!}$ provide$_{\!}$ literature$_{\!}$ review$_{\!}$ and$_{\!}$ related$_{\!}$ discussions$_{\!}$ in$_{\!}$ \S\ref{sec:RW}.

$_{\!}$In$_{\!}$ \S\ref{sec:main-result},$_{\!}$ we$_{\!}$ show$_{\!}$ our$_{\!}$ method$_{\!}$ achieves$_{\!}$ impressive$_{\!}$~results$_{\!}$ over$_{\!}$ famous$_{\!}$ datasets$_{\!}$ (\ie,$_{\!}$ ADE20K$_{\!}$~\cite{zhou2017scene},$_{\!}$ Cityscapes$_{\!}$~\cite{cordts2016cityscapes},$_{\!}$ COCO-Stuff$_{\!}$~\cite{caesar2018coco})$_{\!}$ with$_{\!}$ top-leading$_{\!}$ FCN based$_{\!}$ and$_{\!}$~attention $_{\!\!\!}$ based$_{\!}$ segmentation$_{\!}$ models$_{\!}$ (\ie,$_{\!}$ HRNet$_{\!}$~\cite{wang2020deep}, Swin~\cite{liu2021swin},$_{\!}$ SegFormer$_{\!}$~\cite{xie2021segformer})$_{\!}$ and$_{\!}$ backbones$_{\!}$ (\ie,$_{\!}$ ResNet$_{\!}$~\cite{he2016deep},$_{\!}$
HRNet \cite{wang2020deep}, {Swin$_{\!}$~\cite{liu2021swin}}, MiT$_{\!}$~\cite{xie2021segformer}). Compared with the paramet-\\
\noindent  ric counterparts, our method does not cause any extra computational overhead during testing while reduces the
amount$_{\!}$ of$_{\!}$ learnable$_{\!}$ parameters.$_{\!}$ In$_{\!}$ \S\ref{sec:scale},$_{\!}$ we$_{\!}$ demonstrate$_{\!}$ our$_{\!}$ method$_{\!}$ consistently performs well when increasing the number of semantic classes from 150 to 847.
Accompanied with a set of ablative studies in \S\ref{sec:ablation}, our extensive experiments verify the power of our idea and the efficacy of our algorithm.

Finally, we draw conclusions in \S\ref{sec:conclusion}. This work is expec- ted to open a new venue for future research in$_{\!}$ this$_{\!}$ field. 
\vspace{-2pt}
\section{Existing Semantic Segmentation Models as Parametric Prototype Learning}\label{sec:rev}
Next we first formalize the existing two mask decoding strategies mentioned in \S\ref{sec:intro}, and then answer question~\myhyperlink{Q1}{\ding{182}} from a unified view of parametric prototype learning.

\noindent\textbf{Parametric Softmax Projection}. Almost all FCN-like and many attention-based segmentation models adopt this strategy. Their models comprise two learnable parts: i) an encoder $\phi$ for dense visual feature extraction, and ii) a classifier $\rho$ (\ie, projection head) that projects pixel features into the semantic label space. For each pixel example $i$, its embedding $\bm{i}\!\in\!\mathbb{R}^D$, extracted from $\phi$, is fed into $\rho$ for $C$-way classification:
\vspace{-5pt}
\begin{equation}\small
\begin{aligned}\label{eq:PSP}
p(c|\bm{i})=\frac{\exp(\bm{w}_{c}^\top\bm{i})}{\textstyle\sum^C_{c'=1}\exp(\bm{w}_{c'}^\top \bm{i})},
\end{aligned}
\vspace{-2pt}
\end{equation}
where $p(c|\bm{i})\!\in_{\!}\![0,1]$ is the probability that $i$ being assigned to class $c$. $\rho$ is a pixel-wise linear layer, parameterized by $\bm{W}_{\!}\!=\![\bm{w}_1,\cdots_{\!},\bm{w}_C]\!\in\!\mathbb{R}^{C\!\times_{\!}D\!}$;  $\bm{w}_c\!\in\!\mathbb{R}^{D}$ is a learnable projection vector for $c$-th class; the bias term is omitted for brevity.

\noindent\textbf{Parametric Pixel-Query}. A few attention-based segmentation networks \cite{zheng2021rethinking,xie2021segformer} work in a more `Transformer-like' manner: given the pixel embedding $\bm{i}\!\in\!\mathbb{R}^D$, a set of $C$ query vectors, \ie, $\bm{E}_{\!}\!=\![\bm{e}_1,\cdots_{\!},\bm{e}_C]\!\in\!\mathbb{R}^{C_{\!}\times_{\!}D\!}$, are learned to  generate a probability distribution over the $C$ classes:
\vspace{-3pt}
\begin{equation}\small
\begin{aligned}\label{eq:PPQ}
p(c|\bm{i})=\frac{\exp(\bm{e}_{c\!}\ast\bm{i})}{\textstyle\sum^C_{c'=1}\exp(\bm{e}_{c'\!}\ast\bm{i})},
\end{aligned}
\vspace{-2pt}
\end{equation}
where `$\ast$' is inner product between $\ell_2$-normalized inputs.

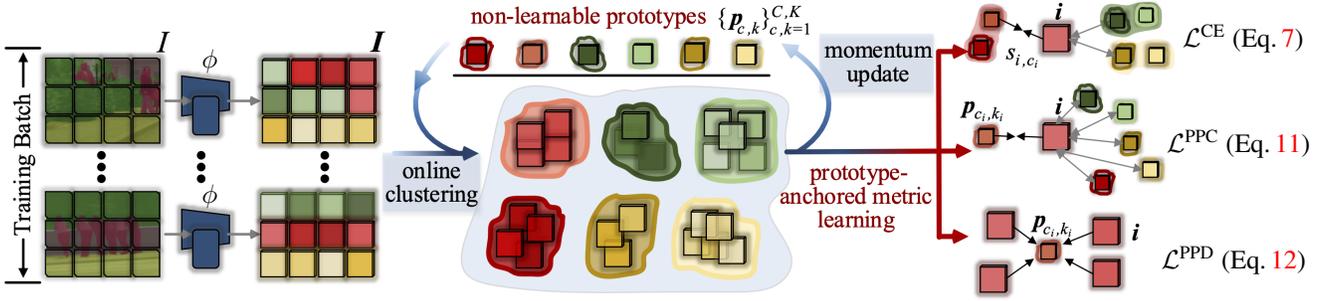
\begin{figure*}[t]
	\vspace{-4pt}
	\begin{center}
		\includegraphics[width=\linewidth]{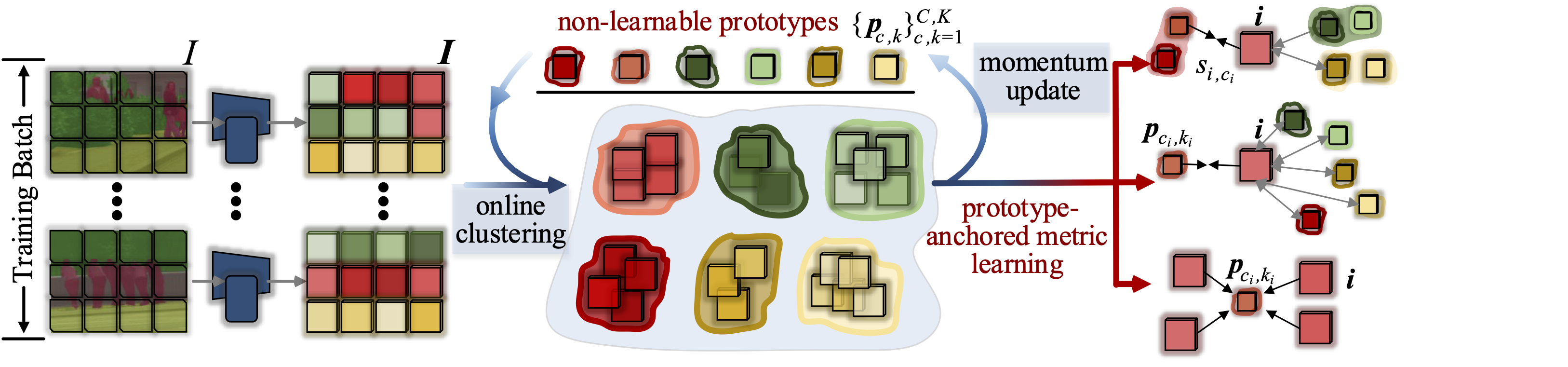}
		\put(-51,98){$\mathcal{L}^{\text{CE}}$ (Eq.$_{\!}$~\ref{eq:nce})}
		\put(-57,58){$\mathcal{L}^{\text{PPC}}$ (Eq.$_{\!}$~\ref{eq:ppc})}
		\put(-59,14){$\mathcal{L}^{\text{PPD}}$ (Eq.$_{\!}$~\ref{eq:ppd})}
		\put(-422,90){$\phi$}
		\put(-422,39){$\phi$}
	\end{center}
	\vspace{-16pt}
	\captionsetup{font=small} \caption{\small\textbf{Architecture illustration} of our non-learnable prototype based nonparametric segmentation model during the training phase.}
	\vspace{-10pt}
	\label{fig:framework}
\end{figure*}

\noindent\textbf{Prototype-based Classification}. Prototype-based classification~\cite{duda1973pattern,friedman2009elements} has been studied for a long time, dating back to the nearest neighbors algorithm$_{\!}$~\cite{cover1967nearest} in machine learning and \textit{prototype theory}$_{\!}$~\cite{rosch1973natural,knowlton1993learning} in cognitive science. Its prevalence stems from its intuitive idea: represent classes by prototypes, and refer to prototypes for classification. Let $\{p_m\}_{m=1\!}^{M}$ be a set of prototypes that are representative of their corresponding classes $\{c_{p_m\!\!}\!\in_{\!}\!\{1, \cdots_{\!}, C\}\}_{m}$. For~a
data sample $i$, prediction is made by
comparing $i$ with $\{p_m\}_{m}$, and taking the class of the \textit{winning} prototype as response:
\vspace{-3pt}
\begin{equation}\small
\begin{aligned}\label{eq:PBC}
\hat{c}_i = c_{p_{m^{\!*}}}, ~~~\text{with}~~~m^* = \mathop{\arg\min}\limits_m\{\langle\bm{i}, \bm{p}_m\rangle\}_{m=1}^M,
\end{aligned}
\vspace{-5pt}
\end{equation}
where $\bm{i}$ and $\{\bm{p}_m\}_{m\!}$ are embeddings of the data sample and prototypes in a feature space, and $\langle\cdot,\cdot\rangle$ stands for the distance measure, which is typically set as $\ell_2$ distance (\ie, $||\bm{i}\!-\!\bm{p}_{m}||$)~\cite{yang2018robust}, yet other proximities can be applied.

Further, Eqs.$_{\!}$~\ref{eq:PSP}-\ref{eq:PPQ} can be formulated in a unified form:
\vspace{-3pt}
\begin{equation}\small
\begin{aligned}\label{eq:PPL}
p(c|\bm{i})=\frac{\exp(-\langle\bm{i}, \bm{g}_c\rangle)}{\sum^C_{c'=1}\exp(-\langle\bm{i}, \bm{g}_{c'}\rangle)},
\end{aligned}
\vspace{-0pt}
\end{equation}
where $\bm{g}_{c}\!\in\!\mathbb{R}^D$ can be either $\bm{w}_{c}$ in Eq.~\ref{eq:PSP} or $\bm{e}_{c}$ in Eq.~\ref{eq:PPQ}.

$_{\!}$With$_{\!}$ Eqs.$_{\!}$~\ref{eq:PBC}-\ref{eq:PPL},$_{\!}$ we$_{\!}$ are$_{\!}$ ready$_{\!}$ to$_{\!}$ answer$_{\!}$ questions$_{\!}$ \myhyperlink{Q1}{\ding{182}}\myhyperlink{Q2}{\ding{183}}.$_{\!}$~Both the two types of methods are based on \textit{learnable} prototypes; they are \textit{parametric} models in the sense that they learn one prototype $\bm{g}_c$, \ie, linear weight $\bm{w}_{c}$ or query vector $\bm{e}_{c}$, for each$_{\!}$  class$_{\!}$  $c$$_{\!}$  (\ie,$_{\!}$  $M_{\!}\!=_{\!}\!C\!~$).$_{\!}$  Thus$_{\!}$  one$_{\!}$  can$_{\!}$  consider$_{\!}$  softmax$_{\!}$~pro- jection based methods `secretly' learn the query vectors. As\\
\noindent  for the difference, in addition to different distance measures (\ie, inner product \textit{vs} cosine similarity), pixel-query based methods~\cite{zheng2021rethinking,xie2021segformer} can feed the queries into cross-attention decoder layers for cross-class context exchanging, rather than softmax projection based counterparts only leveraging the learned class weights within the softmax layer.

%

With the unified view of parametric prototype learning, a few intrinsic yet long ignored issues in this field unfold:

First, \textit{prototype selection}~\cite{garcia2012prototype} is a vital aspect in the design of a
prototype based learner  -- prototypes should be \textit{typical} for their classes. Nevertheless, existing semantic segmentation algorithms often describe each class by only one prototype, bearing no intra-class variation. Moreover, the prototypes are directly learned in a fully parametric manner, without accounting for their representative ability.


$_{\!}$Second,$_{\!}$ the$_{\!}$ amount$_{\!}$ of$_{\!}$ the$_{\!}$ learnable$_{\!}$ prototype$_{\!}$ parameters, \ie,$_{\!}$ $\{\bm{g}_{c\!}\!\in_{\!}\!\mathbb{R}^D\}_{c=1}^C$,
grows with the number of classes. This may hinder the scalability, especially when a large number of classes are present. For example,~if there are 800 classes and the pixel feature dimensionality is 512, at least 0.4M parameters are needed for prototype learning \textit{alone}, making large-vocabulary segmentation a hard task. Moreover, if we want to represent each class by ten prototypes, instead of only one,  we need to learn 4M prototype parameters.


$_{\!}$Third,$_{\!}$ Eq.$_{\!}$~\ref{eq:PBC}$_{\!}$ intuitively$_{\!}$ shows$_{\!}$ that$_{\!}$ prototype based$_{\!}$ learners make metric comparisons of data$_{\!}$~\cite{biehl2013distance}. However, existing algorithms often supervise dense segmentation representation by directly optimizing the accuracy of pixel-wise prediction (\eg, cross-entropy loss), ignoring known inductive biases$_{\!}$~\cite{mitchell1980need,mettes2019hyperspherical}, \eg, intra-class compactness, about the feature distribution. This will hinder the discrimination potential of the learned segmentation features,
as suggested by many literature in representation learning $_{\!}$~\cite{liu2016large,schroff2015facenet,wen2016discriminative}.

$_{\!}$After$_{\!}$ tackling$_{\!}$ question$_{\!}$~\myhyperlink{Q3}{\ding{184}},$_{\!}$ in$_{\!}$ the$_{\!}$ next$_{\!}$ section$_{\!}$ we$_{\!}$~will$_{\!}$ detail$_{\!}$\\
\noindent our$_{\!}$ non-learnable$_{\!}$ prototype$_{\!}$ based$_{\!}$ nonparametric$_{\!}$ segmenta- tion method, which serves as a solid response to question$_{\!}$~\myhyperlink{Q4}{\ding{185}}.

\vspace{-4pt}
\section{Non-Learnable Prototype based Nonparametric Semantic Segmentation}\label{sec:method}
\vspace{-2pt}
We build a nonparametric segmentation framework that$_{\!}$ conducts$_{\!}$ dense$_{\!}$ prediction$_{\!}$ by$_{\!}$ a$_{\!}$ set$_{\!}$ of$_{\!}$ non-learnable$_{\!}$ class$_{\!}$ pro- totypes,$_{\!}$ and$_{\!}$ directly$_{\!}$ supervises$_{\!}$ the$_{\!}$ pixel$_{\!}$ embedding$_{\!}$ space$_{\!}$ via a {prototype-anchored$_{\!}$ metric$_{\!}$ learning} scheme  (Fig.$_{\!}$~\ref{fig:framework}).

%

\noindent\textbf{Non-Learnable$_{\!}$ Prototype$_{\!}$ based$_{\!}$ Pixel$_{\!}$ Classification.}$_{\!}$~As normal,$_{\!}$ an$_{\!}$ encoder$_{\!}$ network$_{\!}$ (FCN$_{\!}$ based$_{\!}$ or$_{\!}$ attention~based), \ie,  $\phi$, is first adopted to map the input image $I_{\!}\!\in_{\!}\!\mathbb{R}^{h_{\!}\times_{\!}w_{\!}\times_{\!}3\!}$,
to a 3D feature tensor $\bm{I}_{\!}\!\in_{\!}\!\mathbb{R}^{H_{\!}\times_{\!}W_{\!}\times_{\!}D\!}$. For pixel-wise $C$-way classification, rather than prior semantic segmentation models that automatically learn $C$ class weights $\{\bm{w}_{c\!}\!\in_{\!}\!\mathbb{R}^{D\!}\}_{c=1\!}^C$ (\textit{cf}.$_{\!}$~Eq.$_{\!}$~\ref{eq:PSP}) or $C$ queries vectors $\{\bm{e}_{c\!}\!\in_{\!}\!\mathbb{R}^{D\!}\}_{c=1\!}^C$ (\textit{cf}.$_{\!}$~Eq.$_{\!}$~\ref{eq:PPQ}), we refer to a group of $CK$~non-learnable prototypes, \ie,  $\{\bm{p}_{c,k\!}\!\in\!\mathbb{R}^{D\!}\}_{c,k=1}^{C,K}$, which$_{\!}$ are$_{\!}$~based$_{\!}$ solely$_{\!}$~on class data sub-centers. More specifically, each class $c\!\in\!\{1, \cdots_{\!}, C\}$ is represented by a total of $K$ prototypes $\{\bm{p}_{c,k\!}\}_{k=1}^K$, and prototype $\bm{p}_{c,k}$ is determined as the center of $k$-th sub-cluster of training pixel samples belonging to  class $c$, in the embedding  space $\phi$. In this way, the prototypes can comprehensively capture characteristic properties of the corresponding classes, without introducing extra learnable parameters  outside $\phi$. Analogous to Eq.$_{\!}$~\ref{eq:PBC}, the category prediction of each pixel $i\!\in\!I$ is achieved by a winner-take-all classification:
\vspace{-2pt}
\begin{equation}\small
\begin{aligned}\label{eq:NIF}
\hat{c}_i = c^{*}, ~~\text{with}~~(c^*,k^*) = \mathop{\arg\min}\limits_{(c,k)}\{\langle\bm{i}, \bm{p}_{c,k}\rangle\}_{c,k=1}^{C,K},
\end{aligned}
\vspace{-4pt}
\end{equation}
where $\bm{i}\!\in\!\mathbb{R}^{D\!}$ stands for the $\ell_2$-normalized embedding of pixel $i$, \ie, $\bm{i}\!\in_{\!}\!\bm{I}$, and the distance measure$_{\!}$ $\langle\cdot,\cdot\rangle$$_{\!}$ is defined as the negative cosine similarity, \ie,$_{\!}$ $\langle \bm{i}, \bm{p} \rangle\!=\!-\bm{i}^{\top\!}\bm{p}$.

With this exemplar-based reasoning mode, we first define the probability distribution of pixel $i$ over the $C$ classes:
\vspace{-7pt}
\begin{equation}\small
\begin{aligned}\label{eq:np}
\!\!\!\!\!p(c|\bm{i})\!=\!\frac{\exp(-s_{i,c})}{\sum^C_{c'=1\!}\exp(-s_{i,c'\!})}, ~~\text{with}~~ s_{i,c} \!=\! \min \{ \langle \bm{i}, \bm{p}_{c,k} \rangle \}_{k=1}^K,\!\!
\end{aligned}
\vspace{3pt}
\end{equation}
where the \textit{pixel-class distance} $s_{i,c}\!\in_{\!}\![-1,1]$ is computed as the distance to the closest prototype of class $c$. Given the groundtruth class of each pixel $i$,~\ie, $c_{i}\!\in\!\{1, \cdots_{\!}, C\}$, the cross-entropy loss can be therefore used for training:
\vspace{-2pt}
\begin{equation}\small
\begin{aligned}\label{eq:nce}
\mathcal{L}_{\text{CE}} &= -\log p(c_i|\bm{i}) \\[-2pt]
&=  -\log \frac{\exp(-s_{i,c_i})}{\exp(-s_{i,c_i})\!+\!\sum_{c'\!\neq c_i\!}\exp(-s_{i,c'})}.\!\!
\end{aligned}
\vspace{-0pt}
\end{equation}
In our case, Eq.$_{\!}$~\ref{eq:nce} can be viewed as pushing pixel $i$ closer to the nearest prototype of its corresponding class, \ie, $c_i$, and further from other close prototypes of irrelevant classes, \ie, $c'\!\neq c_i$. However, only adopting such training objective is not enough, due to two reasons. First, Eq.$_{\!}$~\ref{eq:nce} only considers

\noindent pixel-class distances, \eg, $s_{i,c}$, without addressing within-class pixel-prototype relations, \eg, $\langle \bm{i}, \bm{p}_{c_i,k} \rangle$. For example,

\noindent for discriminative representation learning, pixel $i$ is expec- ted to be pushed further close to a certain prototype (\ie, a particularly suitable pattern) of class $c_i$, and, distant from other prototypes (\ie, other irrelevant but within-class patterns) of class $c_i$. Eq.$_{\!}$~\ref{eq:nce} cannot capture this nature. Second,

\noindent as the pixel-class distances are normalized across all~classes

\noindent (\textit{cf}.$_{\!}$~Eq.$_{\!}$~\ref{eq:np}),$_{\!}$ Eq.$_{\!}$~\ref{eq:nce}$_{\!}$ only$_{\!}$ optimizes$_{\!}$ the$_{\!}$ \textit{relative}$_{\!}$ relations$_{\!}$ between$_{\!}$

\noindent intra-class (\ie, $s_{i,c_i}$) and inter-class (\ie, $\{s_{i,c'\!}\}_{c'\!\neq c_i}$) distances, instead of directly regularizing the cosine distances between pixels and classes. For example, when the intra-class distance $s_{i,c_i}$ of pixel $i$ is relatively smaller than other inter-class distances $\{s_{i,c'}\}_{c'\!\neq c_i}$, the penalty from Eq.$_{\!}$~\ref{eq:nce} will be small, but the intra-class distance $s_{i,c_i\!}$ might still be large \cite{zhang2020rbf,pang2020rethinking}. Next we first elaborate on our within-class online clustering strategy and then detail our two extra training objectives which rely on prototype assignments (\ie, clustering results) and address the above two issues respectively.

\noindent\textbf{Within-Class Online Clustering.}  We approach online~clu- stering for prototype selection and assignment: pixel samples within the same class are assigned to the prototypes belonging to that class, and the
prototypes are then updated according to the assignments. Clustering imposes a natural bottleneck~\cite{ji2019invariant} that forces the model to discover intra-class discriminative patterns yet discard instance-specific details. Thus the prototypes, selected as the sub-cluster centers, are typical of the corresponding classes. Conducting clustering \textit{online} makes our method scalable to large amounts of data, instead of offline clustering requiring multiple passes over the entire dataset for feature computation~\cite{caron2020unsupervised}.



Formally, given pixels $\mathcal{I}^c\!=\!\{i_n\}_{n=1\!}^N$ in a training batch that belong to class $c$ (\ie,$_{\!}$ $c_{i_n\!}\!=\!c$), our goal is to map the pixels $\mathcal{I}^c$ to the $K$ prototypes $\{\bm{p}_{c,k\!}\}_{k=1\!}^K$ of class $c$. We denote this pixel-to-prototype mapping as $\bm{L}^{c\!}\!=\![\bm{l}_{i_n}]_{n=1\!}^N\!\in\!\{0,1\}^{K \times N}$, where $\bm{l}_{i_n\!}\!=\![l_{i_n,k}]_{k=1\!}^K\!\in\!\{0,1\}^{K}$ is the one-hot assignment vector of pixel $i_n$ over the $K$ prototypes. The optimization of $\bm{L}^{c\!}$ is achieved by maximizing the similarity between pixel embeddings, \ie, $\bm{X}^{c\!}\!=\![\bm{i}_{n}]_{n=1\!}^N\!\in\!\mathbb{R}^{D\times N}$, and the prototypes, \ie, $\bm{P}^{c\!}\!=\![\bm{p}_{c,k}]_{k=1\!}^K\!\in\!\mathbb{R}^{D\times K}$:
\vspace{-0pt}
\begin{equation}\small
\begin{aligned}\label{eq:nc1}
&\!\!\!\!\!\!\!\!\max_{\bm{L}^{c}}\texttt{Tr}(\bm{L}^{c\top\!}\bm{P}^{c\top\!}\bm{X}^c), \\[-4pt]
\textit{s.t.}~~~~\bm{L}^{c\!}\!\in\!\{0,1\}^{K \times N}, &~\bm{L}^{c\top}\bm{1}^K=\bm{1}^N, \bm{L}^{c}\bm{1}^N=\frac{N}{K}\bm{1}^K,
\end{aligned}
\vspace{-4pt}
\end{equation}
where $\bm{1}^{K\!}$ denotes the vector of all ones of  $K$ dimensions. The unique assignment constraint, \ie, $\bm{L}^{c\top}\bm{1}^{K\!}\!=_{\!}\!\bm{1}^N$, ensures that each pixel is assigned to one and only one prototype. The  equipartition constraint, \ie, $\bm{L}^{c}\bm{1}^{N\!}\!=\!\frac{N}{K}\bm{1}^K$, enforces that on average each prototype is selected at least $\frac{N}{K}$ times in the batch$_{\!}$~\cite{caron2020unsupervised}. This prevents the trivial solution: all pixel samples are assigned to a single prototype, and eventually benefits  the representative ability of the prototypes. To solve Eq.$_{\!}$~\ref{eq:nc1},  one can relax $\bm{L}^{c}$ to be an element of the \textit{transportation polytope}$_{\!}$~\cite{cuturi2013sinkhorn,asano2020self}:
\vspace{1pt}
\begin{equation}\small
\begin{aligned}\label{eq:nc2}
&\!\!\!\!\!\!\!\!\!\!\!\!\!\!\!\!\max_{{\bm{L}}^{c}}\texttt{Tr}({\bm{L}}^{c\top\!}\bm{P}^{c\top\!}\bm{X}^c)+\kappa\!~h(\bm{L}^{c}), \\[-4pt]
\textit{s.t.}~~~~{\bm{L}}^{c\!}\!\in\!\mathbb{R}^{K \times N}_+&, ~{\bm{L}}^{c\top}\bm{1}^K=\bm{1}^N, {\bm{L}}^{c}\bm{1}^N=\frac{N}{K}\bm{1}^K,
\end{aligned}
\vspace{-2pt}
\end{equation}
where $h(\bm{L}^{c})_{\!}\!=_{\!}\!\sum_{n,k\!}-l_{i_n,k}\log l_{i_n,k\!}$ is an entropy, and $\kappa\!>\!0$ is a parameter that controls the smoothness of distribution. With the soft assignment relaxation and the extra regularization term $h(\bm{L}^{c})$, the solver of Eq.~\ref{eq:nc2} can be given as$_{\!}$~\cite{cuturi2013sinkhorn}:
\vspace{-3pt}
\begin{equation}\small
	\begin{aligned}\label{eq:nc3}
		{\bm{L}}^{c}  = \texttt{diag}(\bm{u}) \exp \big(\frac{\bm{P}^{c\top\!}\bm{X}^c}{\kappa}\big)\texttt{diag}(\bm{v}),
	\end{aligned}
\vspace{-0pt}
\end{equation}
where $\bm{u}\!\in\!\mathbb{R}^{K\!}$ and $\bm{v}\!\in\!\mathbb{R}^{N\!}$ are renormalization vectors, computed by few steps of \textit{Sinkhorn-Knopp iteration}~\cite{cuturi2013sinkhorn}. Our online clustering is highly efficient on GPU, as it only involves a couple of matrix multiplications; in practice, clustering $10$K pixels into $10$ prototypes takes only $2.5$ ms.

\noindent\textbf{Pixel-Prototype Contrastive Learning.} With the assign- ment probability matrix ${\bm{L}}^{c\!}\!=\![{\bm{l}}_{i_n}]_{n=1\!}^N\!\in\![0,1]^{K \times N}$,~we online group the training pixels $\mathcal{I}^{c\!}\!=\!\{i_n\}_{n=1\!}^N$ into $K$ prototypes $\{\bm{p}_{c,k\!}\}_{k=1\!}^K$  within class $c$.  After all the samples in current batch are processed, each pixel $i$ is assigned to $k_{i}$-th prototype of class $c_{i}$, where $k_{i}\!=\!\arg\max_{k}\{{l}_{i,k}\}_{k=1}^K$ and ${l}_{i,k}\!\in\!{\bm{l}}_{i}$. It is natural to derive a training objective for prototype assignment prediction, \ie, maximize the prototype assignment posterior probability. This can be viewed as a pixel-prototype contrastive learning strategy, and addresses

\noindent the first limitation of Eq.~\ref{eq:nce}:
\vspace{-2pt}
\begin{equation}\small
\begin{aligned}\label{eq:ppc}
\!\!\!\mathcal{L}_{\text{PPC}}\!=\!-\!\log \frac{\exp(\bm{i}^\top\!\bm{p}_{c_i,k_i}/\tau)}{\exp(\bm{i}^\top\!\bm{p}_{c_i,k_i\!}/\tau)\!+\!\sum_{\bm{p}^-\in\mathcal{P}^-\!} \exp(\bm{i}^\top\!\bm{p}^-/\tau)},\!\!
\end{aligned}
\vspace{-2pt}
\end{equation}
where $\mathcal{P}^{-\!}\!=\!\{\bm{p}_{c,k}\}_{c,k=1}^{C,K}\big/\bm{p}_{c_i,k_i}$, and the temperature $\tau$ controls the concentration level of representations. Intuitively, Eq.~\ref{eq:ppc} enforces each pixel embedding $\bm{i}$ to be similar with its$_{\!}$ assigned$_{\!}$ (`positive')$_{\!}$ prototype$_{\!}$ $\bm{p}_{c_i,k_i}$,$_{\!}$ and$_{\!}$ dissimilar$_{\!}$ with other $CK\!-\!1$ irrelevant (`negative') prototypes $\mathcal{P}^{-}$. Compared with prior pixel-wise metric learning based segmentation models~\cite{wang2021exploring}, which consume numerous negative pixel samples,
our method only needs $CK$ prototypes for pixel-prototype contrast computation, \textit{neither} causing large memory cost \textit{nor} requiring heavy pixel pair-wise comparison. 


\noindent\textbf{Pixel-Prototype$_{\!}$ Distance$_{\!}$ Optimization.}$_{\!}$ Building$_{\!}$ upon$_{\!}$~the relative comparison over pixel-class/-prototype distances, Eq.$_{\!}$~\ref{eq:nce}$_{\!}$ and Eq.$_{\!}$~\ref{eq:ppc} inspire inter-class/-cluster discrinimitive- ness, but less consider reducing the intra-cluster variation, \ie, making pixel features of the same prototype compact. Thus a compactness-aware loss is used for further regulari- zing representations by directly minimizing the distance be- tween$_{\!}$ each$_{\!}$ embedded$_{\!}$ pixel$_{\!}$ and$_{\!}$ its$_{\!}$ assigned$_{\!}$ prototype:
\vspace{-2pt}
\begin{equation}\small
\begin{aligned}\label{eq:ppd}
\!\!\!\mathcal{L}_{\text{PPD}}\!=\!(1-\bm{i}^{\top\!}\bm{p}_{c_i,k_i})^2.
\end{aligned}
\vspace{1pt}
\end{equation}
Note that both $\bm{i}$ and $\bm{p}_{c_i,k_i\!}$ are $\ell_2$-normalized. This training objective minimizes intra-cluster variations while maintaining separation between features with different prototype assignments, making our model more robust against outliers.



\begin{figure}[t]
	\vspace{-4pt}
	\begin{center}
		\includegraphics[width=\linewidth]{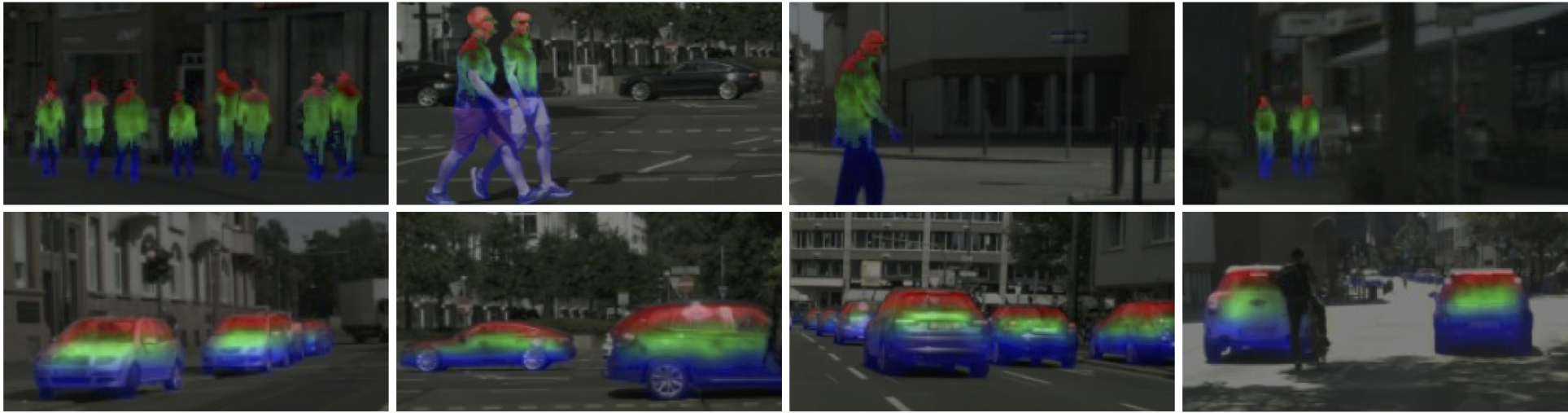}
	\end{center}
	\vspace{-18pt}
	\captionsetup{font=small}
	\caption{\small \textbf{Visualization of pixel-prototype similarity} for \textit{person} (top) and \textit{car} (bottom) classes. Please refer to \S\ref{sec:method} for details. }
	\vspace{-15pt}
	\label{fig:prototype}
\end{figure}

\noindent\textbf{Network Learning and Prototype Update.} Our model is a nonparametric approach that learns semantic segmentation by directly optimizing the pixel embedding space $\phi$. It is called nonparametric because it constructs prototype hypotheses directly from the training pixel samples themselves. Thus the parameters of the feature extractor $\phi$ are learned through stochastic gradient descent, by minimizing the combinatorial loss over all the training pixel samples:
\vspace{-1.3pt}
\begin{equation}\small
\begin{aligned}\label{eq:com}
\mathcal{L}_{\text{SEG}}=\mathcal{L}_{\text{CE}} + \lambda_1\mathcal{L}_{\text{PPC}}+ \lambda_2\mathcal{L}_{\text{PPD}}.
\end{aligned}
\vspace{-2.3pt}
\end{equation}
Meanwhile, the non-learnable prototypes $\{\bm{p}_{c,k}\}_{c,k=1\!}^{C,K}$ are \textit{not} learned by stochastic gradient descent, but are computed as the centers of the corresponding embedded pixel samples. To do so, we let the prototypes evolve continuously by accounting for the online clustering results. Particularly, after each training iteration, each prototype is updated as:
\vspace{-2.3pt}
\begin{equation}\small
\begin{aligned}\label{eq:update}
\bm{p}_{c,k} \leftarrow  \mu\bm{p}_{c,k} + (1-\mu)\bar{\bm{i}}_{c,k},
\end{aligned}
\vspace{-1.3pt}
\end{equation}
where $\mu\!\in\![0,1]$ is a momentum coefficient, and $\bar{\bm{i}}_{c,k}$ indicates the $\ell_2$-normalized, mean vector of the embedded training pixels, which are assigned to prototype $\bm{p}_{c,k}$ by online clustering. With the clear meaning of the prototypes, our segmentation procedure can be intuitively understood as retrieving the most similar prototypes (sub-class centers). {Fig.~\ref{fig:prototype} provides prototype retrieval results for \textit{person} and \textit{car}
with $K\!=\!3$ prototypes for each. The prototypes are associ-

\noindent ated with different colors (\ie, red, green, and blue). For each pixel, its distance to the closest prototype is visualized using the corresponding prototype color. As can be seen, the prototypes well correspond to meaningful patterns within classes, validating their representativeness. }

\vspace{-5pt}
\section{Related Work}\label{sec:RW}
\vspace{-3pt}
In this section, we review representative work in semantic segmentation, prototype learning and metric learning.

\noindent\textbf{Semantic Segmentation.} Recent years have witnessed remarkable progress in semantic segmentation, due to$_{\!}$ the$_{\!}$~fast evolution of backbone architectures $_{\!}$--$_{\!}$ from$_{\!}$ CNN-based$_{\!}$ (\eg,$_{\!}$ VGG$_{\!}$~\cite{Simonyan15},$_{\!}$ ResNet$_{\!}$~\cite{he2016deep})$_{\!}$ to$_{\!}$ Transformer-like$_{\!}$~\cite{vaswani2017attention}$_{\!}$ (\eg,$_{\!}$ ViT\\
\noindent \cite{dosovitskiy2020image}, Swin$_{\!}$~\cite{liu2021swin}), and$_{\!}$ segmentation$_{\!}$ models $_{\!}$--$_{\!}$ from FCNs \cite{long2015fully} to attention networks (\eg,$_{\!}$ SegFormer$_{\!}$~\cite{xie2021segformer}).$_{\!}$  Specifically,$_{\!}$ FCN$_{\!}$~\cite{long2015fully}$_{\!}$~is$_{\!}$ a$_{\!}$ milestone;$_{\!}$ it learns dense prediction efficiently.\\
\noindent  Since it was proposed, numerous efforts have~been devoted\\
\noindent   to improving FCN, by, for example,  enlarging~the receptive$_{\!}$ field \cite{zhao2017pyramid,dai2017deformable,yang2018denseaspp,yu2015multi,chen2017deeplab,chen2018encoder};  strengthening context~cues \cite{ronneberger2015u,zheng2015conditional,yu2015multi,badrinarayanan2017segnet,liu2017deep,lin2017refinenet,mehta2018espnet,zhang2018context,he2019adaptive,hu2020class,yuan2020object,yu2020context,liu2021exploit,Hsiao_2021_ICCV,Jin_2021_ICCV2,Jin_2021_ICCV,zhou2021group}; leveraging boundary information \cite{chen2016semantic,yu2018learning,ding2019boundary,bertasius2016semantic,li2020improving,yuan2020segfix,zhen2020joint};$_{\!}$ incorporating$_{\!}$  neural$_{\!}$ attention \cite{harley2017segmentation,wang2018non,li2018pyramid,zhao2018psanet,hu2018squeeze,wang2021hierarchical,he2019dynamic,li2019expectation,fu2019dual,huang2019ccnet,sun2020mining}; or automating network engineering
 \cite{chen2019fasterseg,liu2019auto,nekrasov2019fast,li2020learning}. Lately,$_{\!}$ Transfor- mer based$_{\!}$ solutions$_{\!}$ \cite{xie2021segformer,strudel2021segmenter,zheng2021rethinking,cheng2021maskformer} attained growing attention; enjoying the flexibility in long-range dependency modeling, fully attentive solutions yield impressive results.

Different from current  approaches that are typically built upon$_{\!}$ learnable$_{\!}$ prototypes,$_{\!}$ in$_{\!}$~pre-deep$_{\!}$ era,$_{\!}$ many$_{\!}$ segmenta- tion systems
are nonparametric \cite{malisiewicz2008recognition,tighe2010superparsing,torralba200880,liu2010sift,liu2011nonparametric,eigen2012nonparametric}. By absorbing their case-based reasoning ideas, we build a nonparametric segmentation network, which explicitly derives prototypes from sample clusters and hence directly optimizes the embedding space with distance metric constraints. In \cite{kong2018recurrent,wang2021exploring}, while cluster-/pixel-level metric loss is adopted to regularize representation, the pixel class is still inferred via parametric softmax. \cite{jain2021scaling} purely relies on class\\
\noindent embeddings,$_{\!}$ which,$_{\!}$ however,$_{\!}$ are$_{\!}$ fully$_{\!}$ trainable.$_{\!}$ Thus$_{\!}$~\cite{kong2018recurrent,wang2021exploring,jain2021scaling} are all parametric methods. As far as we know,~\cite{hwang2019segsort}\\
\noindent is$_{\!}$ the$_{\!}$ only$_{\!}$ non-learnable$_{\!}$ prototype,$_{\!}$ deep$_{\!}$ learning$_{\!}$ based$_{\!}$~se- mantic$_{\!}$ segmentation$_{\!}$ model.$_{\!}$ But$_{\!}$ \cite{hwang2019segsort} treats image regions as prototypes, incurring huge memory and computational demand. Besides, \cite{hwang2019segsort} only considers the relative difference between inter-and intra-class sample-prototype distances like the parametric counterparts. Our method is more principled with fewer heuristic$_{\!}$ designs.$_{\!}$ Unlike$_{\!}$~\cite{hwang2019segsort},$_{\!}$ we represent prototypes as sub-cluster centers and obtain online assignments, allowing our method to scale gracefully to~any\\
\noindent  dataset size. We encourage a sparse distance distribution with compactness-awareness, reinforcing the embedding discrimination.$_{\!}$ With$_{\!}$ a$_{\!}$ broader$_{\!}$ view,$_{\!}$ a$_{\!}$ few$_{\!}$ embedding based instance segmentation approaches \cite{de2017semantic,neven2019instance} can be viewed as nonparametric, \ie, treat instance centroids as prototypes.

\noindent\textbf{Prototype Learning.} Cognitive psychological studies evi- dence that people use past cases as models when learning to\\
\noindent solve problems$_{\!}$~\cite{newell1972human,aamodt1994case,yang2021multiple}. Among various machine learning algorithms, ranging from classical statistics based methods to Support Vector Machine to Multilayer Perceptrons$_{\!}$~\cite{duda1973pattern,shawe2004kernel,friedman2009elements,bishop2006pattern},  prototype based classification gains particular interest, due to its exemplar-driven nature and intuitive interpretation:$_{\!}$ observations$_{\!}$ are$_{\!}$ directly$_{\!}$ compared$_{\!}$ with re-
presentative examples. Based on the nearest neighbors~rule -- the earliest prototype learning method$_{\!}$~\cite{cover1967nearest}, many famous,
nonparametric classifiers are proposed$_{\!}$~\cite{garcia2012prototype}, such as Learning Vector Quantization (LVQ) \cite{kohonen1998self},  generalized LVQ$_{\!}$~\cite{sato1995generalized}, and Neighborhood Component Analysis$_{\!}$~\cite{goldberger2004neighbourhood,salakhutdinov2007learning}. There~has been$_{\!}$ a$_{\!}$ recent$_{\!}$ surge$_{\!}$ of$_{\!}$
interest$_{\!}$ to$_{\!}$ integrate$_{\!}$ deep$_{\!}$ learning$_{\!}$~into prototype$_{\!}$ learning,$_{\!}$ showing$_{\!}$ good$_{\!}$ potential$_{\!}$ in$_{\!}$ few-shot \cite{snell2017prototypical}, zero-shot$_{\!}$~\cite{jetley2015prototypical},$_{\!}$ and$_{\!}$ unsupervised$_{\!}$ learning$_{\!}$~\cite{wu2018unsupervised,xu2020attribute},$_{\!}$ as$_{\!}$ well$_{\!}$ as$_{\!}$ supervised$_{\!}$
classification$_{\!}$~\cite{wu2018improving,yang2018robust,guerriero2018deepncm,mettes2019hyperspherical}$_{\!}$ and$_{\!}$ interpretable$_{\!}$ networks$_{\!}$  \cite{li2018deep}.$_{\!}$  Remarkably,$_{\!}$  as$_{\!}$  many$_{\!}$  few-shot segmentation models can be viewed as prototype-based networks \cite{vinyals2016matching,dong2018few,wang2019panet},  our work sheds light on the possibility of closer collaboration between the two segmentation fields.



\noindent\textbf{Metric Learning.}  The selection of proper distance measure$_{\!}$ impacts$_{\!}$
the$_{\!}$ success of prototype based learners$_{\!}$~\cite{biehl2013distance}; metric learning  and prototype learning are naturally related. As~the
literature on metric learning is vast~\cite{kaya2019deep}, only the most relevant ones are discussed. The goal of metric learning is to learn a distance metric/embedding such that similar samples are pulled together and dissimilar samples are pushed away. It has shown a significant benefit by learning deep representation using metric loss functions (\eg, contrastive loss \cite{hadsell2006dimensionality}, triplet loss \cite{schroff2015facenet}, $n$-pair loss$_{\!}$ \cite{sohn2016improved}) for applications (\eg, image retrieval \cite{wang2014learning}, face recognition \cite{schroff2015facenet}). Recently, metric\\
\noindent learning showed good potential in unsupervised representa- tion learning.$_{\!}$ Specifically,$_{\!}$ many$_{\!}$ \textit{instance-based}$_{\!}$ approaches
use the contrastive loss \cite{gutmann2010noise,oord2018representation} to  explicitly compare pairs of image representations, so as to push away features
from different images while pulling together those from transformations  of the same image \cite{oord2018representation,hjelm2019learning,chen2020improved,chen2020simple,he2020momentum}. Since computing all the pairwise comparisons on a large dataset is challenging, some \textit{clustering-based} methods turn to discriminate
between groups of images with similar features instead of individual images \cite{cliquecnn2016,xie2016unsupervised,caron2018deep,yan2020clusterfit,caron2020unsupervised,li2020prototypical,asano2020self,van2020scan,tao2021idfd}. Our prototype-anchored metric learning strategy shares a similar spirit of posing metric constraints over prototype (cluster) assignments, but it is to reshape the pixel segmentation embedding space with explicit supervision.

\vspace{-2pt}
\section{Experiment}\label{sec:exp}
\vspace{-3pt}
\subsection{Experimental Setup}
\vspace{-3pt}
\noindent\textbf{Datasets.}~Our experiments are conducted on three datasets:\!\!
\vspace{1pt}
\begin{itemize}[leftmargin=*]
	\setlength{\itemsep}{0pt}
	\setlength{\parsep}{-2pt}
	\setlength{\parskip}{-0pt}
	\setlength{\leftmargin}{-10pt}
	\vspace{-6pt}
	
	\item \textbf{ADE20K}~\cite{zhou2017scene} is a large-scale scene parsing benchmark that covers $150$ stuff/object categories. The dataset is divided into 20k/2k/3k images for \texttt{train}/\texttt{val}/\texttt{test}.
		
	\item \textbf{Cityscapes}~\cite{cordts2016cityscapes} has 5k  finely annotated urban scene images, with $2,\!975/500/1,\!524$ for \texttt{train}/\texttt{val}/\texttt{test}. The segmentation performance is evaluated over $19$  challenging categories, such as rider, bicycle, and traffic light.
	
	\item \textbf{COCO-Stuff}~\cite{caesar2018coco} has 10k images gathered from COCO \cite{lin2014microsoft}, with 9k and 1k for \texttt{train} and \texttt{test}, respectively. There are 172 semantic categories in total, including 80 objects, 91 stuffs and 1 unlabeled.
	
	\vspace{-4pt}
\end{itemize}

\noindent\textbf{Training.}
Our method is implemented on  {MMSegmentation}~\cite{mmseg2020}, following default training settings. In particular, all backbones are initialized using corresponding weights pre-trained on ImageNet-1K~\cite{ImageNet}, while remaining layers are randomly initialized. We use standard data augmentation techniques, including random scale jittering with a factor in  $[0.5, 2]$, random horizontal flipping, random cropping as well as random color jittering.  We train models using SGD/AdamW for FCN-/attention-based models, respectively. The learning rate is scheduled  following the polynomial annealing policy.  In addition,  for Cityscapes, we use a batch size of $8$, and a training crop size of $768\!\times\!768$. For ADE20K and COCO-Stuff, we use a crop size of $512\!\times\!512$ and train the models with batch size $16$. The models are trained for $160$k, $160$k, and $80$k iterations on Cityscapes, ADE20K and COCO-Stuff, respectively. Exceptionally, for ablation study, we train models for 40K iterations. The hyper-parameters are empirically set to: $K\!=\!10$, $m\!=\!0.999$, $\tau\!=\!0.1$, $\kappa\!=\!0.05$, $\lambda_1\!=\!0.01$, $\lambda_2\!=\!0.01$.

\noindent\textbf{Testing.} For ADE20K and COCO-Stuff, we rescale the short scale of the image to training crop size, with the aspect ratio kept unchanged. For Cityscapes, we adopt sliding window inference with the window size  $768\times768$. For simplicity, we do not apply \textit{any} test-time data augmentation. Our model is implemented in PyTorch and trained on eight  Tesla V100 GPUs with a 32GB memory per-card. Testing is conducted on the same machine.

\noindent\textbf{Baselines.} We mainly compare with four widely recognized segmentation models, \ie, two FCN based (\ie, FCN~\cite{long2015fully}, HRNet~\cite{wang2020deep}) and two attention based (\ie, Swin~\cite{liu2021swin} and SegFormer~\cite{xie2021segformer}).  For fair comparison,  all the models  are based on our reproduction, following the hyper-parameter and augmentation recipes used in {MMSegmentation}~\cite{mmseg2020}.



\noindent\textbf{Evaluation Metric.}  Following conventions~\cite{long2015fully,chen2017deeplab}, mean intersection-over-union (mIoU) is adopted for evaluation.

\newcommand{\reshl}[2]{
\textbf{#1} \fontsize{7.5pt}{1em}\selectfont\color{mygreen}{$\uparrow$ \textbf{#2}}
}
		
\begin{table}[t]
	\small
\resizebox{\columnwidth}{!}{
	\tablestyle{4pt}{1.02}
	\begin{tabular}{|r|r||cc|}
		\thickhline
		\rowcolor{mygray}
		\multicolumn{1}{|c|}{Method} &  \multicolumn{1}{c||}{Backbone}   & \tabincell{c}{\# Param\\(M)} &  \tabincell{c}{~~mIoU~~\\(\%)}  \\
		
		\hline\hline
		DeepLabV3+~\pub{ECCV18}~\cite{chen2018encoder}
		& ResNet-101~\cite{he2016deep} & 62.7 &  {44.1}  \\
		OCR~\pub{ECCV20}~\cite{yuan2020object}
		& HRNetV2-W48~\cite{wang2020deep} & 70.3 &   {45.6}  \\
		MaskFormer~\pub{NeurIPS21}~\cite{cheng2021maskformer}
		& ResNet-101~\cite{he2016deep} & 60.0 &   {46.0}  \\
		UperNet~\pub{ECCV20}~\cite{xiao2018unified}
		& Swin-Base~\cite{liu2021swin} & 121.0 & 48.4 \\
		OCR~\pub{ECCV20}~\cite{yuan2020object}
		& HRFormer-B~\cite{YuanFHLZCW21} & 70.3 &   {48.7}  \\
		SETR~\pub{CVPR21}~\cite{zheng2021rethinking}
		& ViT-Large~\cite{dosovitskiy2020image}  & 318.3 & 50.2 \\
		Segmenter~\pub{ICCV21}~\cite{strudel2021segmenter}
		& ViT-Large~\cite{dosovitskiy2020image} & 334.0 &  {51.8}  \\
		$^\dagger$MaskFormer~\pub{NeurIPS21}~\cite{cheng2021maskformer}
		& Swin-Base~\cite{liu2021swin} & 102.0 &   {52.7}  \\
		\hline
		
		FCN~\pub{CVPR15}~\cite{long2015fully}
		&  & 68.6 & 39.9  \\
		
		\textbf{\texttt{Ours}}
		& \multirow{-2}{*}{ResNet-101~\cite{he2016deep}} & 68.5 & \reshl{41.1}{1.2} \\
		
		HRNet~\pub{PAMI20}~\cite{wang2020deep}
		&  & 65.9 & 42.0  \\
		
		\textbf{\texttt{Ours}}
		& \multirow{-2}{*}{HRNetV2-W48~\cite{wang2020deep}} & 65.8 &  \reshl{43.0}{1.0} \\
		\hline
		
		Swin~\pub{ICCV21}~\cite{liu2021swin}
		&   & 90.6 &  {48.0} \\
		\textbf{\texttt{Ours}}
		& \multirow{-2}{*}{Swin-Base~\cite{liu2021swin}}   & 90.5 & \reshl{48.6}{0.6}  \\

		SegFormer~\pub{NeurIPS21}~\cite{xie2021segformer}
		&  & 64.1 & {50.9}  \\ 
        \textbf{\texttt{Ours}}
		& \multirow{-2}{*}{MiT-B4~\cite{xie2021segformer}} &  64.0  &  \reshl{51.7}{0.8}  \\ 
		
		\hline
	\end{tabular}
}
\leftline{~\footnotesize{$^\dagger$: backbone is pre-trained on ImageNet-22K.}}
		\vspace{-12pt}
\captionsetup{font=small}
	\caption{\small\textbf{Quantitative results} (\S\ref{sec:main-result}) on ADE20K$_{\!}$~\cite{zhou2017scene} \texttt{val}.}
	\vspace{-10pt}
	\label{table:ade}
\end{table}

\renewcommand{\hl}[1]{\textcolor{Highlight}{#1}}
\definecolor{Highlight}{HTML}{39b54a}  

\begin{table}[t]
	\small
\resizebox{\columnwidth}{!}{
	\tablestyle{4.5pt}{1.02}
	\begin{tabular}{|r|r||cc|}
		\thickhline
		\rowcolor{mygray}
		\multicolumn{1}{|c|}{Method} &  \multicolumn{1}{c||}{Backbone}   & \tabincell{c}{\# Param\\(M)} &  \tabincell{c}{~~mIoU~~\\(\%)}  \\
		
		\hline\hline
		
		PSPNet~\pub{CVPR17}~\cite{zhao2017pyramid}
		& ResNet-101~\cite{he2016deep} & 65.9 & 78.4 \\
		PSANet~\pub{ECCV18}~\cite{zhao2018psanet}
		& ResNet-101~\cite{he2016deep} & -& 78.6 \\
		AAF~\pub{ECCV18}~\cite{ke2018adaptive}
		& ResNet-101~\cite{he2016deep} & -& 79.1 \\
	
		Segmenter~\pub{ICCV21}~\cite{strudel2021segmenter}
		& ViT-Large~\cite{dosovitskiy2020image} & 322.0 & {79.1}  \\
		
		ContrastiveSeg~\pub{ICCV21}~\cite{wang2021exploring}
		& ResNet-101~\cite{he2016deep} & 58.0  & 79.2 \\

		MaskFormer~\pub{NeurIPS21}~\cite{cheng2021maskformer}
		& ResNet-101~\cite{he2016deep} & 60.0 &   {80.3}  \\
		DeepLabV3+~\pub{ECCV18}~\cite{chen2018encoder}
		& ResNet-101~\cite{he2016deep} & 62.7 & 80.9 \\
		OCR~\pub{ECCV20}~\cite{yuan2020object}
		& HRNetV2-W48~\cite{wang2020deep} & 70.3  & 81.1 \\

		\hline
		
		FCN~\pub{CVPR15}~\cite{long2015fully}
		&  & 68.6 & 78.1 \\
		
		\textbf{\texttt{Ours}}
		& \multirow{-2}{*}{ResNet-101~\cite{he2016deep}} & 68.5& \reshl{79.1}{1.0} \\
		
		HRNet~\pub{PAMI20}~\cite{wang2020deep}
		&  & 65.9 & 80.4  \\
		
		\textbf{\texttt{Ours}}
		& \multirow{-2}{*}{HRNetV2-W48~\cite{wang2020deep}} &65.8 & \reshl{81.1}{0.7}  \\
		\hline
		
		Swin~\pub{ICCV21}~\cite{liu2021swin}
		&   &  90.6 & 79.8 \\
		\textbf{\texttt{Ours}}
		& \multirow{-2}{*}{Swin-Base~\cite{liu2021swin}}   & 90.5 & \reshl{80.6}{0.8}    \\
		SegFormer~\pub{NeurIPS21}~\cite{xie2021segformer}
		&   & 64.1  & {80.7} \\ 
		\textbf{\texttt{Ours}}
		& \multirow{-2}{*}{MiT-B4~\cite{xie2021segformer}}  &  64.0& \reshl{81.3}{0.6}    \\ 
		
		\hline
	\end{tabular}}
	\vspace{-3pt}
	\captionsetup{font=small}
	\caption{\small\textbf{Quantitative results} (\S\ref{sec:main-result}) on Cityscapes \cite{cordts2016cityscapes} \texttt{val}.}
	\vspace{-16pt}
	\label{table:city}
\end{table}


%
%
%
%
%
%
%
%
%
%
%
%
%

\begin{figure*}[t]
	\vspace{-2pt}
	\begin{center}
		\includegraphics[width=\linewidth]{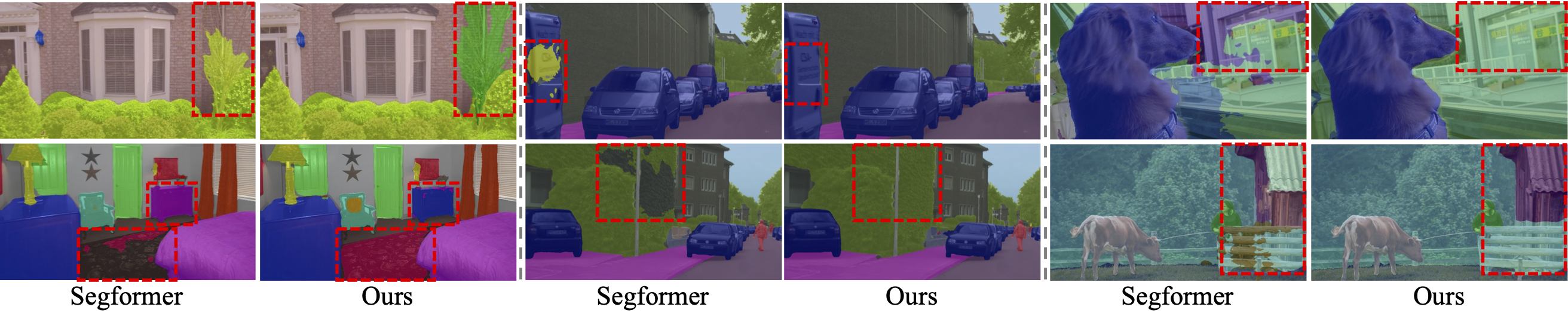}
	\end{center}
	\vspace{-20pt}
	\captionsetup{font=small}
	\caption{\small\textbf{Qualitative results} of Segformer~\cite{xie2021segformer} and our approach (from left to right: ADE20K~\cite{zhou2017scene}, Cityscapes~\cite{cordts2016cityscapes}, COCO-Stuff~\cite{caesar2018coco}).}
	\vspace{-15pt}
	\label{fig:visual}
\end{figure*}

	\vspace{-2pt}
\subsection{Comparison to State-of-the-Arts}\label{sec:main-result}
	\vspace{-1pt}
\noindent\textbf{ADE20K~\cite{zhou2017scene} \texttt{val}.}
Table~\ref{table:ade} reports comparisons with representative models on ADE20K \texttt{val}. Our nonparametric scheme obtains consistent improvements over the baselines, with fewer learnable parameters. In particular, it yields \textbf{1.2\%} and \textbf{1.0\%} mIoU improvements over the FCN-based counterparts, \ie, FCN~\cite{long2015fully} and HRNet~\cite{wang2020deep}. Similar performance gains (\textbf{0.6\%} and \textbf{0.8\%}) are obtained over recent attention-based models, \ie, Swin~\cite{liu2021swin} and SegFormer \cite{xie2021segformer}, manifesting the high versatility of our approach.

\noindent\textbf{Cityscapes~\cite{cordts2016cityscapes} \texttt{val}.} Table~\ref{table:city} shows again our compelling performance on Cityscapes \texttt{val}. Specifically, our approach surpasses all the competitors, \ie, \textbf{1.0\%} over FCN, \textbf{0.7\%} over HRNet, \textbf{0.8\%} over Swin, and \textbf{{0.6\%}} over Segformer.

\noindent\textbf{COCO-Stuff~\cite{caesar2018coco} \texttt{test}.}
As listed in Table~\ref{table:coco}, our approach also demonstrates promising performance on COCO-Stuff \texttt{test}. It outperforms all the baselines. Notably, with MiT-B4~\cite{xie2021segformer} as the network backbone, our approach earns an mIoU score of \textbf{43.3\%}, establishing a new state-of-the-art.


\noindent\textbf{Qualitative Results.}
Fig.~\ref{fig:visual} provides qualitative comparison of \textbf{\texttt{Ours}} against Segformer~\cite{xie2021segformer} on representative examples in the three datasets. We observe that our approach is able to handle diverse challenging scenarios and produce more accurate results (as highlighted in red dashed boxes).

\begin{table}
	\centering
	\small
\resizebox{\columnwidth}{!}{
	\tablestyle{4.5pt}{1.02}
	\begin{tabular}{|r|r||cc|}
		\thickhline
		\rowcolor{mygray}
		\multicolumn{1}{|c|}{Method} &  \multicolumn{1}{c||}{Backbone}   & \tabincell{c}{\# Param\\(M)} &  \tabincell{c}{~~mIoU~~\\(\%)}  \\ \hline\hline
		
		SVCNet~\pub{CVPR19}~\cite{ding2019semantic}
		& ResNet-101~\cite{he2016deep} & -& 39.6 \\
		DANet~\pub{CVPR19}~\cite{fu2019dual}
		& ResNet-101~\cite{he2016deep} & 69.1 & 39.7  \\
		SpyGR~\pub{CVPR20}~\cite{li2020spatial}
		& ResNet-101~\cite{he2016deep} &- & 39.9 \\
		MaskFormer~\pub{NeurIPS21}~\cite{cheng2021maskformer}
		& ResNet-101~\cite{he2016deep} & 60.0 &   {39.8}  \\
		ACNet~\pub{ICCV19}~\cite{fu2019adaptive}
		& ResNet-101~\cite{he2016deep} & -& 40.1 \\
		OCR~\pub{ECCV20}~\cite{yuan2020object}
		& HRNetV2-W48~\cite{wang2020deep} & 70.3 & 40.5   \\
		\hline
		
		FCN~\pub{CVPR15}~\cite{long2015fully}
		&  & 68.6 &  32.5 \\
		
		\textbf{\texttt{Ours}}
		& \multirow{-2}{*}{ResNet-101~\cite{he2016deep}} &68.5 & \reshl{34.0}{1.5}  \\
		
		HRNet~\pub{PAMI21}~\cite{wang2020deep}
		&  & 65.9 & 38.7 \\
		\textbf{\texttt{Ours}}
		& \multirow{-2}{*}{HRNetV2-W48~\cite{wang2020deep}} & 65.8& \reshl{39.9}{1.2}  \\ \hline
		
		
		Swin~\pub{ICCV21}~\cite{liu2021swin}
		&   & 90.6  & 41.5 \\
	    \textbf{\texttt{Ours}}
	    & \multirow{-2}{*}{Swin-Base~\cite{liu2021swin}}   &90.5 &  \reshl{42.4}{0.9}   \\
		SegFormer~\pub{NeurIPS21}~\cite{xie2021segformer}
		&   & 64.1 &  42.5  \\
		\textbf{\texttt{Ours}}
		& \multirow{-2}{*}{MiT-B4~\cite{xie2021segformer}}  & 64.0&  \reshl{43.3}{0.8}    \\
		\hline
	\end{tabular}}
	\vspace{-3pt}
	\captionsetup{font=small}
	\caption{\small\textbf{Quantitative results} (\S\ref{sec:main-result})$_{\!}$ on COCO-Stuff~\cite{caesar2018coco} \texttt{test}.}
	\label{table:coco}
	\vspace{-16pt}
\end{table}

\begin{table*}[t]
	\small
	\vspace{-3pt}
	\resizebox{\textwidth}{!}{
		\tablestyle{1.5pt}{1.02}
		\begin{tabular}{|c|c||cc|cc|cc|cc|cc|}
			\thickhline
			\rowcolor{mygray}
			&  & \multicolumn{2}{c|}{150 classes} & \multicolumn{2}{c|}{300 classes} & \multicolumn{2}{c|}{500 classes} & \multicolumn{2}{c|}{700 classes}& \multicolumn{2}{c|}{847 classes}  \\
			\rowcolor{mygray}
			\multirow{-2}{*}{Method}  & \multirow{-2}{*}{\# Proto}
			& mIoU (\%) & \# Param (M)  & mIoU (\%) & \# Param (M)   & mIoU (\%) & \# Param (M) & mIoU (\%) & \# Param (M) & mIoU (\%) & \# Param (M) \\
			\hline\hline
			
			parametric & 1
			& {45.1} & 27.48 (0.12) 			 &  36.5  & 27.62 (0.23) 			& 25.7 & 27.80 (0.39) & 19.8  & 27.98 (0.54) & 16.5 & 28.11 (0.65) \\
			\tabincell{c}{nonparametric \\(\textbf{\texttt{Ours}})} & 1
			& \reshl{45.5}{0.4} & 27.37 (0)  & \reshl{37.2}{0.7} &27.37 (0) & \reshl{26.8}{1.1} &27.37 (0)  & \reshl{21.2}{1.4} &27.37 (0)& \reshl{18.1}{1.6} &27.37 (0)\\
			
			\hline
			
			parametric & 10
			& 45.7  & 28.56 (1.2) & 37.0  & 29.66 (2.3) & 26.6  & 31.26 (3.9) & 20.8  & 32.86 (5.4) & 17.7 & 33.96 (6.5) \\
			\tabincell{c}{nonparametric \\(\textbf{\texttt{Ours}})} & 10
			& \reshl{46.4}{0.7}& 27.37 (0)  & \reshl{37.8}{0.8} &27.37 (0) & \reshl{27.9}{1.3} &27.37 (0)  &  \reshl{22.1}{1.3} &27.37 (0)&  \reshl{19.4}{1.7} &27.37 (0)\\
			
			\hline
	\end{tabular}}
	\vspace{-3pt}
	\captionsetup{font=small}
	\caption{\small\textbf{Scalability study} (\S\ref{sec:scale}) of our nonparametric model against the parametric baseline (\ie, SegFormer~\cite{xie2021segformer}) on ADE20K~\cite{zhou2017scene}. For each model variant, we report its segmentation mIoU, parameter numbers of the entire model as well as the prototypes (in the bracket).}
	\vspace{-10pt}
	\label{table:scalebility}
\end{table*}

\begin{table*}[t]
	\subfloat[{Training Objective $\mathcal{L}$} \label{table:loss}]{
		\tablestyle{7pt}{1.05}
		\begin{tabular}{|ccc||c|}\thickhline
			\rowcolor{mygray}
			$\mathcal{L}_{\text{CE}}$ & $\mathcal{L}_{\text{PPC}}$ & $\mathcal{L}_{\text{PPD}}$  & mIoU  \\
			\rowcolor{mygray}
			(Eq.~\ref{eq:nce}) & 	(Eq.~\ref{eq:ppc}) & 	(Eq.~\ref{eq:ppd})  & (\%) \\ \hline\hline
			\cmark & &                    	 & 45.0\\
			\cmark &\cmark &          	&  45.9 \\
			\cmark & &\cmark           	&  45.4\\
			\cmark & \cmark&\cmark & 46.4\\
			\hline
		\end{tabular}
	}\hfill
	\subfloat[{Prototype Number $K$}\label{table:prototypenumber}]{%
		\tablestyle{5pt}{1.05}
		\begin{tabular}{|l||c|}\thickhline
			\rowcolor{mygray}
			\# Prototype & mIoU (\%)\\ \hline\hline
			~~~$K=1$  & 45.5 \\
			~~~$K=5$  & 46.0 \\
			~~~$K=10$  & 46.4 \\
			~~~$K=20$  & 46.5\\
			~~~$K=50$  & 46.4\\
			\hline
		\end{tabular}
	}\hfill
	\subfloat[{Momentum Coefficient $\mu$}\label{table:momentum}]{%
		\tablestyle{6pt}{1.05}
		\begin{tabular}{|l||c|}\thickhline
			\rowcolor{mygray}
			Coefficient $\mu$ & mIoU (\%) \\\hline\hline
			~~$\mu=0$ & 44.9 \\
			~~$\mu=0.9$ & 45.9 \\
			~~$\mu=0.99$ & 46.0 \\
			~~$\mu=0.999$ & 46.4 \\
			~~$\mu=0.9999$ & 46.3 \\
			\hline
		\end{tabular}
	}\hfill
	\subfloat[{Distance Measure}\label{table:distance}]{%
		\tablestyle{8pt}{1.05}
		\begin{tabular}{|c||c|}\thickhline
			\rowcolor{mygray}
			Distance Measure & mIoU (\%) \\ \hline\hline
			Standard & 45.7\\
			Huberized & 45.2 \\
			Cosine  & 46.4 \\
			
			\hline
			\multicolumn{2}{c}{} \\
			\multicolumn{2}{c}{}
		\end{tabular}
	}\hfill
	\vspace{-3pt}
	\captionsetup{font=small}
	\caption{\small A set of \textbf{ablative studies} (\S\ref{sec:ablation}) on ADE20K~\cite{zhou2017scene} \texttt{val}. All model variants use MiT-B2~\cite{xie2021segformer} as the backbone.}
	\label{tab:ablations}\vspace{-5mm}
\end{table*}
	\vspace{-3pt}
\subsection{Scalability to Large-Vocabulary Semantic Segmentation}\label{sec:scale}
	\vspace{-2pt}
Today, rigorous evaluation of semantic segmentation models is mostly performed in a few category regime (\eg, 19/150/172 classes for Cityscapes/ADE20K/COCO-Stuff), while the generalization to  more natural  large-vocabulary setting is ignored. In this section, we demonstrate the remarkable superiority of our method in large-vocabulary setting. We start with the default setting in ADE20K \cite{zhou2017scene} which includes 150 semantic concepts. Then, we gradually increase the number of concepts based on their visibility frequency, and train/test models on the selected number of classes. In this experiment, we use MiT-V2~\cite{xie2021segformer} as the backbone and train models for 40k iterations.

The results are summarized in Table~\ref{table:scalebility}, from which we find that: \textit{i)} For the parametric scheme, the amount of  prototype parameters increases with vocabulary size. For the extreme case of $10$ prototypes and $847$ classes, the number of prototype parameters is $6.5$ M, accounting for  $\sim\!20\%$ of total parameters (\ie, $33.96$ M). In sharp contrast,  our scheme requires no any learnable prototype parameters. \textit{ii)} Our method achieves consistent performance elevations against the parametric counterpart under all settings. These results well demonstrate the utility of our nonparametric  scheme for unrestricted open-vocabulary semantic segmentation.



\subsection{Diagnostic Experiment}\label{sec:ablation}
To investigate the effect of our core designs, we conduct ablative studies on ADE20K$_{\!}$~\cite{zhou2017scene} \texttt{val}. We use MiT-B2 \cite{xie2021segformer} as the backbone and train models for 40K iterations.

\noindent\textbf{Training Objective.}
We first investigate our overall training objective (\textit{cf}.$_{\!}$~Eq.$_{\!}$~\ref{eq:com}). As shown in Table~\ref{table:loss}, the model with $\mathcal{L}_{\text{CE}}$ alone achieves an mIoU score of  $45.0\%$. Adding $\mathcal{L}_{\text{PPC}}$ or $\mathcal{L}_{\text{PPD}}$ individually brings  gains (\ie, $\textbf{0.9\%}$/$\textbf{0.4\%}$), revealing the value to explicitly learn pixel-prototype relations. Combing all the losses together leads to the best performance, yielding an  mIoU score of 46.4\%.


\noindent\textbf{Prototype Number Per Class $K$.}
Table~\ref{table:prototypenumber} reports the performance of our approach with regard to the number of prototype per class. For $K\!=\!1$, we directly represent each class as the mean embedding of its pixel samples. The pixel assignment is based simply on ground-truth labels, without using online clustering (Eqs.~\ref{eq:nc1}-\ref{eq:nc2}). This baseline obtains a score of $45.5\%$.  Further, when using more  prototypes (\ie, $K\!=\!3$), we see a clear performance boost (\ie, $45.5\%\!\rightarrow\!46.0\%$). The  score  further improves when allowing $5$ or $10$ prototypes; however, increasing $K$ beyond $10$ gives marginal returns in performance. As a result, we set $K\!=\!10$ for a better trade-off between accuracy and computation cost. This study confirms our motivation to use multiple prototypes for capturing intra-class variations.

\noindent\textbf{Coefficient $\mu$.}
Table~\ref{table:momentum} quantifies the effect of momentum coefficient ($\mu$ in Eq.~\ref{eq:update}) which controls the speed of prototype updating.  The model performs reasonably well using a relatively large coefficient (\ie, {{$\mu\!\in[0.999,0.9999]$}), showing that a slow updating is beneficial. When {{$\mu$}} is $0.9$ or $0.99$, the performance decreases, and drops considerably at the extreme case of {{$\mu\!=\!0$}}.

\noindent\textbf{Distance Measure.}
By default, we use cosine distance (refer to as `Cosine') to measure pixel-prototype similarity as denoted in Eq.~\ref{eq:np}, Eq.~\ref{eq:ppc} and Eq.~\ref{eq:ppd}. However, other choices are also applicable. Here we study two alternatives.  The first is the standard Euclidean distance (\ie, `Standard'), \ie, $\langle \bm{x}, \bm{y} \rangle\!=\! \|\bm{x}\!-\!\bm{y}\|_2$. In contrast to `Cosine', here $\bm{x}$ and $\bm{y}$ are un-normalized real-valued vectors. To handle the non-differentiability in `Standard', we further study an approximated Huber-like function~\cite{huber1973robust} (`Huberized'), \ie, $\langle \bm{x}, \bm{y} \rangle\!=\!  \delta (\sqrt{\|\bm{x}-\bm{y}\|^2/\delta^2+1}-1)$. The hyper-parameter $\delta$ is empirically set to $0.1$.  As we find from Table~\ref{table:distance} that `Cosine' performs much better than other un-normalized Euclidean measurements. The Huberized norm does not show any advantage over `Standard'.

\vspace{-9pt}
\section{Conclusion and Discussion}\label{sec:conclusion}
\vspace{-3pt}
$_{\!}$The$_{\!}$ vast$_{\!}$ majority$_{\!}$ of$_{\!}$ recent$_{\!}$ effort$_{\!}$ in$_{\!}$ this$_{\!}$ field$_{\!}$ seek$_{\!}$ to$_{\!}$ learn parametric class representations for pixel-wise recognition. In contrast, this paper explores an exemplar-based regime. This leads to a nonparametric segmentation$_{\!}$ framework,$_{\!}$ where$_{\!}$ several$_{\!}$ typical$_{\!}$ points$_{\!}$ in$_{\!}$ the embedding space are selected as class prototypical representation, and distance to the prototypes determines how a pixel sample is classified. It enjoys several advantages: \textbf{i)} explicit prototypical representation for class-level statistics modeling; \textbf{ii}) better generalization with nonparametric pixel-category prediction; and \textbf{iii)} direct optimization of the feature embedding space. Our framework is elegant, general, and yields outstanding performance. It also comes with some intriguing questions. For example, to pursue better interpretability, one can optimize the prototypes to directly resemble pixel- or region-level observations~\cite{li2018deep,hwang2019segsort}. Overall, we feel the results in this
paper warrant further exploration in this direction.


%
%

{
\small
\bibliographystyle{ieee_fullname}
\bibliography{egbib}
}

\clearpage
\appendix

%

\section{Large-Vocabulary Dataset Description}\label{sec:data}

First, we present the details of the datasets used for the study of large-vocabulary  semantic segmentation (\S{\color{red}{5.3}}). In particular,  the study is based on {ADE20K-Full} \cite{zhou2017scene}, which contains 25K and 2K images for \texttt{train} and \texttt{val}, respectively. It is elaborately annotated in an open-vocabulary setting with more than $3,\!000$ semantic concepts. Following \cite{cheng2021maskformer}, we only keep the $847$ concepts that  are appearing in both \texttt{train} and \texttt{val} sets. Then, for each variant ADE20K-$x$ in Table {\color{red}{4}}, we choose the top-$x$ ($x\!\in\!\{300,500,700,847\}$) classes based on their appearing frequencies. Note that for ADE20K-150, we follow the default setting in the SceneParse150 challenge to use the specified $150$ classes for evaluation.

\section{Online Clustering Algorithm}\label{sec:code}

Algorithm~\ref{alg:code} provides a pseudo-code for our online clustering algorithm to solve Eq.~{\color{red}{9}}.  The algorithm only includes a small number of matrix-matrix products, and can run efficiently on a GPU card.

\section{More Experimental Result}\label{sec:exp}

\subsection{Quantitative  Result on Cityscapes \texttt{test}} \label{sec:city}
Table~\ref{table:city} reports the results on Cityscapes \texttt{test}. All the models are trained on \texttt{train}$+$\texttt{val} sets. Note that we do not include any coarsely labeled Cityscapes data for training. For fair comparison with~\cite{xie2021segformer}, we train our model with a cropping size of $1024\!\times\!1024$, and adopt sliding window inference with a window size of $1024\!\times\!1024$. As seen, our approach reaches $\textbf{83.0\%}$ mIoU, which is $\textbf{0.8\%}$ higher than SegFormer~\cite{xie2021segformer}. In addition, it greatly outperforms many famous segmentation models, such as HANet \cite{choi2020cars}, HRNetV2 \cite{wang2020deep}, SETR \cite{zheng2021rethinking}.


\subsection{Quantitative  Result with Lightweight Backbones} \label{sec:backbone}
In Table~\ref{table:lightweight}, we compare our approach against four competitors using lightweight backbones (\ie, MobileNet-V2~\cite{sandler2018mobilenetv2}, MiT-B0~\cite{xie2021segformer}) on ADE20K \cite{zhou2017scene} \texttt{val}. With MiT-B0, our model  achieves the best  performance (\ie, \textbf{38.5\%} mIoU) with the smallest number of parameters (\ie, \textbf{3.7} M).

\subsection{Hyper-parameter Analysis of $\lambda_1$ and $\lambda_2$}

Table~\ref{table:coefficient} summarizes the influence of hyper-parameters $\lambda_1$ and $\lambda_2$ to model performance on ADE20K \cite{zhou2017scene} \texttt{val}. We observe that our model is robust to the two coefficients, and achieves the best performance at $\lambda_1\!=\!0.01,\lambda_2\!=\!0.01$.

\begin{algorithm}[t]
	\caption{Pseudo-code of Online Clustering Algorithm in the PyTorch-like style.}
	\label{alg:code}
	\definecolor{codeblue}{rgb}{0.25,0.5,0.5}
	\algcomment{\fontsize{7.2pt}{0em}\selectfont\texttt{mm}: matrix multiplication. }
	\lstset{
		backgroundcolor=\color{white},
		basicstyle=\fontsize{7.2pt}{7.2pt}\ttfamily\selectfont,
		columns=fullflexible,
		breaklines=true,
		captionpos=b,
		commentstyle=\fontsize{7.2pt}{7.2pt}\color{codeblue},
		keywordstyle=\fontsize{7.2pt}{7.2pt},
	}
\begin{lstlisting}[language=python]
# P: non-learnable prototypes (D x K)
# X: pixel embeddings (D x N)
# iters: sinhorn-knopp iteration number
# kappa: hyper-parameter (Eq.9)
# L: pixel-to-prototype assignment (K x N, Eq.9)
		
def online_clustering(P, X, iters=3, kappa=0.05)
    L = mm(P.transpose(), X)
    L = torch.exp(L / kappa)
    L /= torch.sum(L)
		
    for _ in range(iters):
       # normalize each row
       L /= torch.sum(L, dim=1, keepdim=True)
       L /= K
		
       # normalize each column
       L /= torch.sum(L, dim=0, keepdim=True)
       L /= N
		
    # make sure the sum of each column to be 1
    L *= N
		
    return L
\end{lstlisting}
\end{algorithm}

\begin{table}[t]
	\small
	\resizebox{\columnwidth}{!}{
		\tablestyle{4.5pt}{1.1}
		\begin{tabular}{|r|c||cc|}
			\thickhline
			\rowcolor{mygray}
			\multicolumn{1}{|c|}{Method} &  \multicolumn{1}{c||}{Backbone}   & \tabincell{c}{\# Param\\(M)} &  \tabincell{c}{~~mIoU~~\\(\%)}  \\
			
			\hline\hline
			PSPNet~\pub{CVPR17}~\cite{zhao2017pyramid}
			& ResNet-101~\cite{he2016deep} & 65.9 & 78.4 \\
			PSANet~\pub{ECCV18}~\cite{zhao2018psanet}
			& ResNet-101~\cite{he2016deep} & -& 80.1 \\
			ContrastiveSeg~\pub{ICCV21}~\cite{wang2021exploring}
			& ResNet-101~\cite{he2016deep} & 58.0  & 80.3 \\
			$^\dagger$SETR~\pub{ICCV19}~\cite{zheng2021rethinking}
			& ViT~\cite{dosovitskiy2020image} & 318.3 & 81.0 \\
			HRNetV2~\pub{PAMI20}~\cite{wang2020deep}
			& HRNetV2-W48~\cite{wang2020deep} & 65.9& 81.6 \\
			CCNet~\pub{ICCV19}~\cite{huang2019ccnet}
			& ResNet-101~\cite{he2016deep} & -& 81.9 \\
			HANet~\pub{CVPR20}~\cite{choi2020cars}
			& ResNet-101~\cite{he2016deep} & -& 82.1 \\
			\hline
			
			SegFormer~\pub{NeurIPS21}~\cite{xie2021segformer}
			&   & 84.7  & {82.2} \\ 
			\textbf{\texttt{Ours}}
			& \multirow{-2}{*}{MiT-B5~\cite{xie2021segformer}}  &  84.6 &  \reshl{83.0}{0.8}   \\
			
			\hline
	\end{tabular}}
	\leftline{~\footnotesize{$^\dagger$: backbone is pre-trained on ImageNet-22K.}}
	\captionsetup{font=small}
	\caption{\small\textbf{Quantitative results} (\S\ref{sec:city}) on Cityscapes \cite{cordts2016cityscapes} \texttt{test}.}
	\vspace{-5pt}
	\label{table:city}
\end{table}


\begin{table}[t]
	\small
	\resizebox{\columnwidth}{!}{
		\tablestyle{4.5pt}{1.1}
		\begin{tabular}{|r|c||cc|}
			\thickhline
			\rowcolor{mygray}
			\multicolumn{1}{|c|}{Method} &  \multicolumn{1}{c||}{Backbone}   & \tabincell{c}{\# Param\\(M)} &  \tabincell{c}{~~mIoU~~\\(\%)}  \\
			
			\hline\hline
			
			FCN~\pub{CVPR15}~\cite{long2015fully}
			&  MobileNet-V2~\cite{sandler2018mobilenetv2} & 9.8 & 19.7 \\
			PSPNet~\pub{CVPR17}~\cite{zhao2017pyramid}
			&  MobileNet-V2~\cite{sandler2018mobilenetv2} & 13.7 & 29.6 \\
			DeepLabV3+~\pub{ECCV18}~\cite{chen2018encoder}
			&  MobileNet-V2~\cite{sandler2018mobilenetv2} & 15.4 & 34.0 \\\hline
			SegFormer~\pub{NeurIPS21}~\cite{xie2021segformer}
			&  MiT-B0~\cite{xie2021segformer} & 3.8 & 37.4 \\
			\textbf{\texttt{Ours}}
			&  MiT-B0~\cite{xie2021segformer} & \textbf{3.7} & \reshl{38.5}{1.1} \\
			\hline
	\end{tabular}}
	\captionsetup{font=small}
	\caption{\small\textbf{Quantitative results} on ADE20K \cite{zhou2017scene} \texttt{val} with lightweight backbones. See \S\ref{sec:backbone} for details.}
	\vspace{-7pt}
	\label{table:lightweight}
\end{table}

\begin{table}[h]
	\centering
	\small
	\resizebox{\columnwidth}{!}{
		\setlength\tabcolsep{6pt}
		\renewcommand\arraystretch{1.0}
		\begin{tabular}{r|cccccc}
			\hline
			$\lambda_1$
			& 0.001  & 0.005  & 0.01  & 0.02 & 0.03 & 0.05  \\
			mIoU (\%)
			& 46.1& 46.3 & \textbf{46.4} & \textbf{46.4}  & 46.2  & 46.3 \\ \hline
			$\lambda_2$
			& 0.001  & 0.005  & 0.01  & 0.02 & 0.03 & 0.05   \\
			mIoU (\%)
			& 46.2 &  46.2 & \textbf{46.4}  & 46.3 & 46.3 & 46.1
			\\ \hline
			
		\end{tabular}
	}
	\captionsetup{font=small}
	\caption{\small{Analysis of $\lambda_1$ and $\lambda_2$ on ADE20K \cite{zhou2017scene} \texttt{val}.}}
	\label{table:coefficient}
	
\end{table}

\subsection{Embedding Structure Visualization}
Fig.~\ref{fig:embedding}  visualizes the embedding learned by (left) parametric  \cite{xie2021segformer}, and (right) our nonparametric segmentation model. As seen, in our algorithm, the pixel embeddings belonging to the same prototypes are well separated. This is because that  our model  is essentially based on a distance-based point-wise classifier and its embedding is directly supervised by the metric learning losses, which help reshape the feature space by encoding latent data structure into the embedding space.

\begin{figure}[H]
	\centering
	\begin{subfigure}{0.48\columnwidth}
		\includegraphics[width=\textwidth]{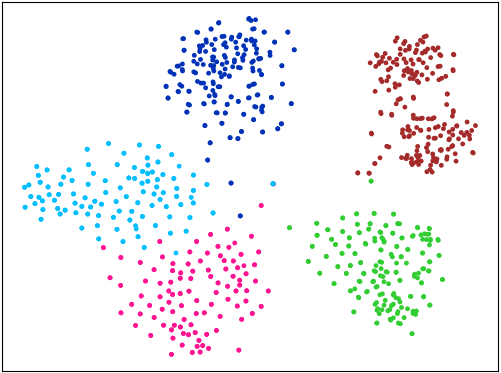}
	\end{subfigure}
	\hspace{0.1cm}
	\begin{subfigure}{0.48\columnwidth}
		\includegraphics[width=\textwidth]{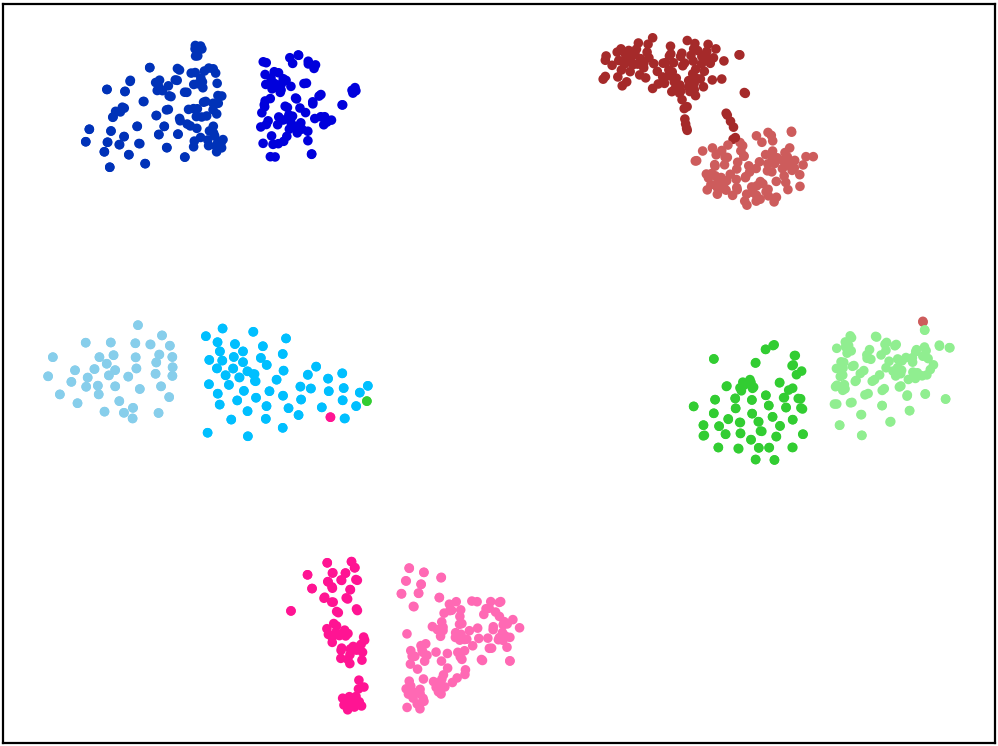}
	\end{subfigure}
	\vspace{-4pt}
	\captionsetup{font=small}
	\caption{\small{\textbf{Embedding spaces} learned by (left) parametric model \cite{xie2021segformer}, and (right) our nonparametric model. For better visualization, we show five classes of Cityscapes \cite{cordts2016cityscapes} with two prototypes per class.}
		\vspace{-5pt}}
	\label{fig:embedding}
\end{figure}

\subsection{Additional Qualitative Result}
We show more qualitative results on ADE20K \cite{zhou2017scene} (Fig.~\ref{fig:ade20k}), Cityscapes \cite{cordts2016cityscapes} (Fig.~\ref{fig:city}) and COCO-Stuff \cite{caesar2018coco} (Fig.~\ref{fig:ooco}). As observed, our approach generally gives more accurate predictions than SegFormer \cite{xie2021segformer}.

\section{Discussion}\label{sec:discusssion}

\noindent\textbf{Limitation Analysis.}
One limitation of our approach is that it needs a clustering procedure during training, which increases the time complexity. However, in practice, the clustering algorithm imposes a minor computational
burden, only taking about 2.5 ms to cluster 10K pixels into 10 prototypes. Additionally, like many other semantic segmentation models, our approach is subject to some factors such as domain gaps, label quality and fairness. We will put more efforts on  improving the ``in the wild'' robustness of our model in the future research.

\noindent\textbf{Broader Impact.}
This work provides a prototype perspective to unify existing mask decoding strategies, and accordingly introduces a novel non-learnable prototype based nonparametric segmentation scheme. On the positive side, the approach pushes the boundary of  segmentation algorithms in terms of model efficiency and accuracy, and shows great potentials in unrestricted segmentation scenarios with  thousands of semantic categories. Thus, the research could find diverse real-world applications such as self-driving cars and robot navigation. On the negative side, our model  can be misused to segment the minority groups for malicious purposes. In addition, the problematic segmentation  may cause inaccurate decision or planning of  systems based on the results.

\noindent\textbf{Future Work.} This work also comes with new challenges, certainly worth further exploration:
\begin{itemize}[leftmargin=*]
	\setlength{\itemsep}{0pt}
	\setlength{\parsep}{1pt}
	\setlength{\parskip}{1pt}
	\setlength{\leftmargin}{-10pt}
	\item \textbf{Closer Ties to Unsupervised Representation Learning.} Our segmentation model directly learns the pixel embedding space with non-learnable prototypes. A critical success factor of recent unsupervised representation learning methods lies on the direct \textit{comparison} of embeddings. By sharing such regime, our nonparametric model has good potential to make full use of unsupervised representations.
	
	\item \textbf{Further Enhancing Interpretability.} Our model only uses the mean of several embedded `support' pixel samples as the prototype for each (sub-)class. To pursue better interpretability, one can optimize the prototypes to directly resemble actual pixels, or region-level observations~\cite{li2018deep,hwang2019segsort}.
	
	\item \textbf{Unifying Image-Wise and Pixel-Wise Classification.} A common practice of building segmentation models is to remove the classification head from a pretrained classifier and leave the encoder. This is not optimal as lots of `knowledge' are directly dropped. However, with prototype learning, one can transfer the `knowledge' of a nonparametric classier to a nonparametric segmenter intactly, and formulate image-wise and pixel-wise classification in a unified paradigm.
\end{itemize}



%

\newpage
\begin{figure*}[t]
	\vspace{-4pt}
	\begin{center}
		\includegraphics[width=\linewidth]{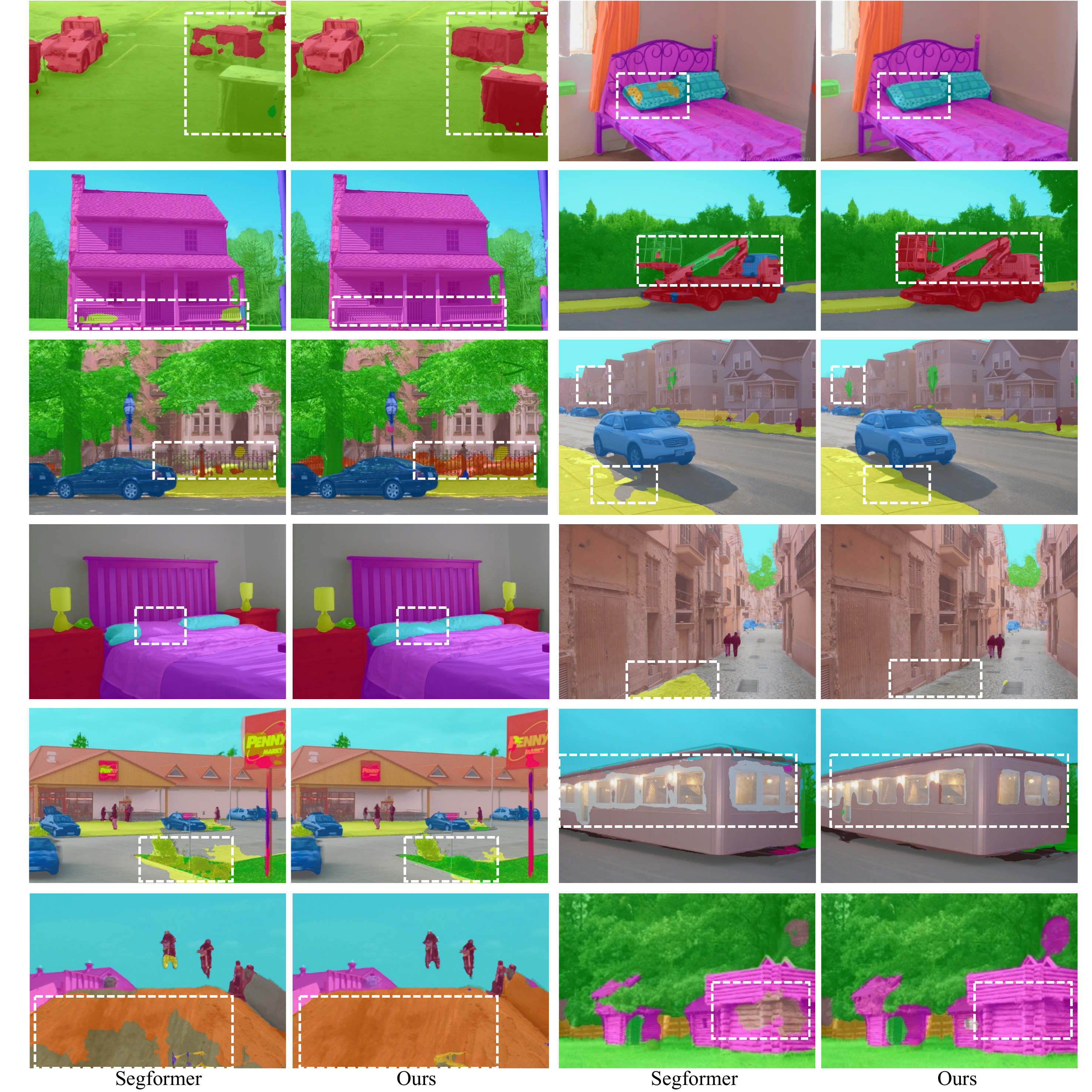}
	\end{center}
	\vspace{-16pt}
	\captionsetup{font=small}
	\caption{\small \textbf{Qualitative results} of Segformer~\cite{xie2021segformer} and our approach on ADE20K~\cite{zhou2017scene} \texttt{val}. }
	\vspace{-10pt}
	\label{fig:ade20k}
\end{figure*}

\begin{figure*}[t]
	\vspace{-4pt}
	\begin{center}
		\includegraphics[width=\linewidth]{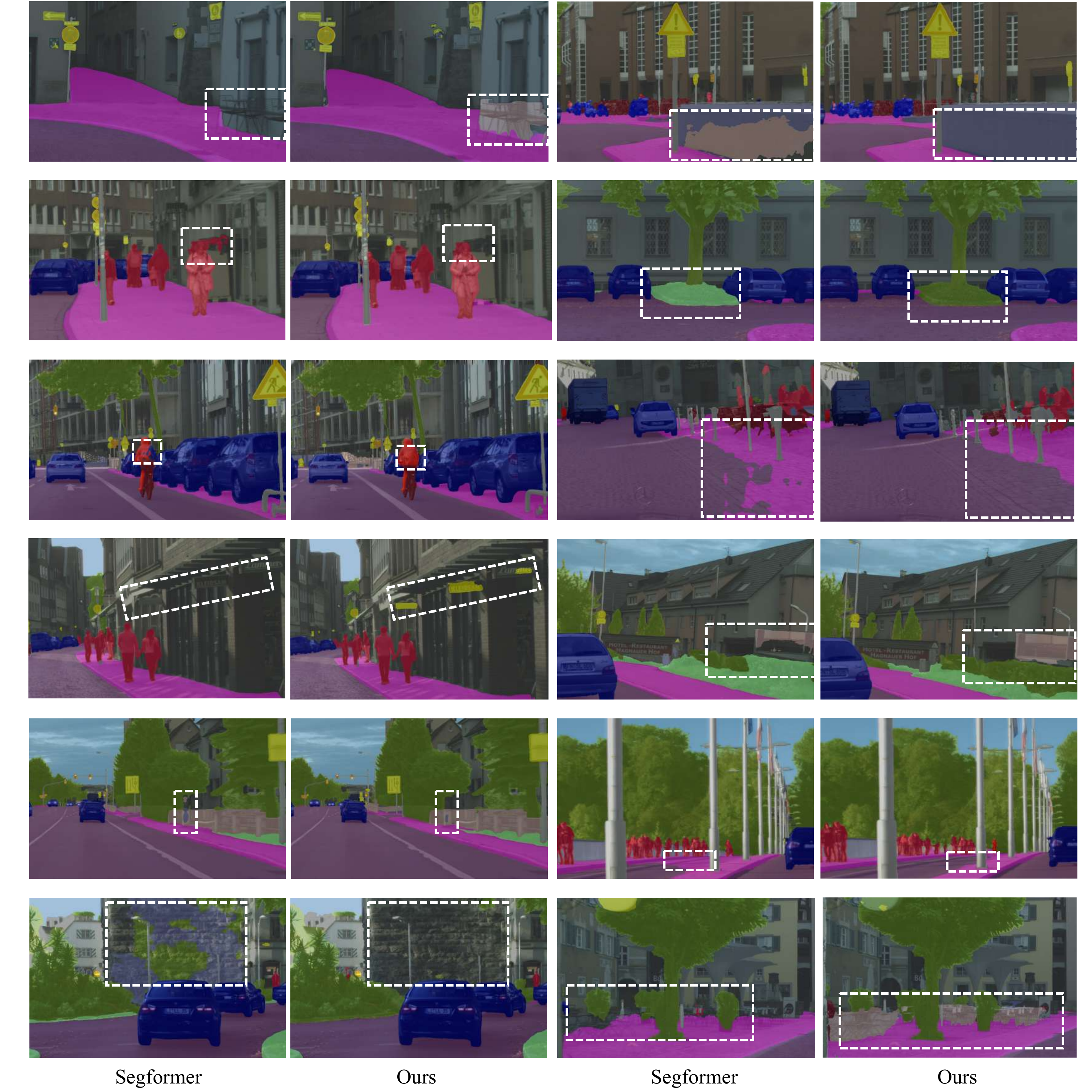}
	\end{center}
	\vspace{-16pt}
	\captionsetup{font=small}
	\caption{\small \textbf{Qualitative results} of Segformer~\cite{xie2021segformer} and our approach on Cityscapes~\cite{cordts2016cityscapes} \texttt{val}. }
	\vspace{-10pt}
	\label{fig:city}
\end{figure*}

\begin{figure*}[t]
	\vspace{-4pt}
	\begin{center}
		\includegraphics[width=\linewidth]{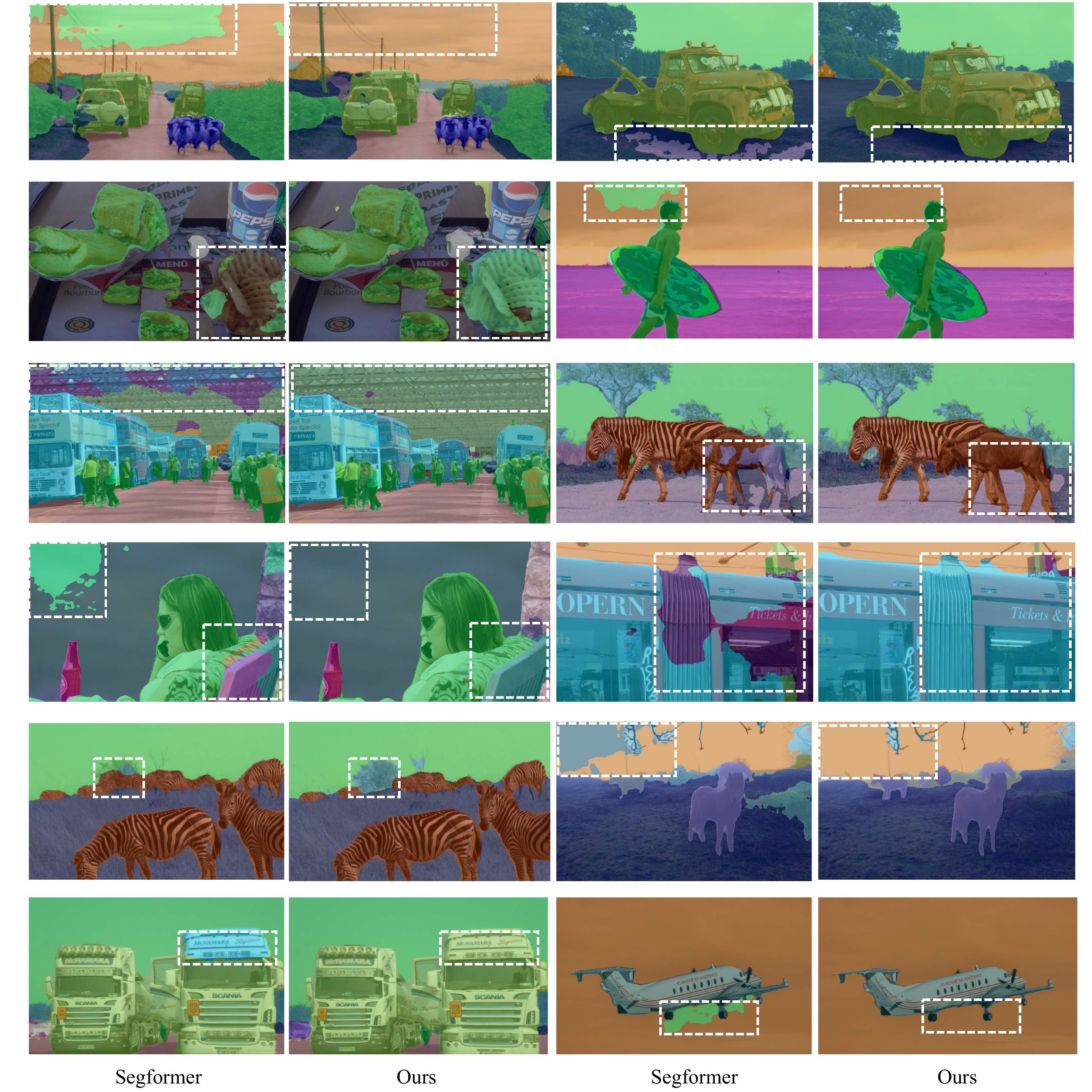}
	\end{center}
	\vspace{-16pt}
	\captionsetup{font=small}
	\caption{\small \textbf{Qualitative results} of Segformer~\cite{xie2021segformer} and our approach on COCO-Stuff~\cite{caesar2018coco} \texttt{test}. }
	\vspace{-10pt}
	\label{fig:ooco}
\end{figure*}

\end{document}